\pdfoutput=1
\documentclass{article}

\usepackage[preprint,nonatbib]{neurips_2022}

\usepackage[utf8]{inputenc} 
\usepackage[T1]{fontenc}    
\usepackage{hyperref}       
\usepackage{url}            
\usepackage{booktabs}       
\usepackage{amsfonts}       
\usepackage{nicefrac}       
\usepackage{microtype}      
\usepackage{xcolor}         
\usepackage{bm}
\usepackage{amsmath}
\usepackage{amssymb}
\usepackage{tabularx}
\usepackage{multirow}
\usepackage{multicol}
\usepackage{pifont}
\usepackage{bigdelim}
\usepackage{rotating}
\usepackage{makecell}
\usepackage[numbers]{natbib}

\title{VD-PCR: Improving Visual Dialog with\\Pronoun Coreference Resolution}

\author{%
  Xintong Yu \\
  Department of Automation\\
  Tsinghua University\\
  Beijing, P.R.China \\
  \texttt{yuxt16@mails.tsinghua.edu.cn} \\
   \And
   Hongming Zhang \\
   Department of CSE \\
   HKUST \\
   Hong Kong, P.R.China \\
   \texttt{hzhangal@cse.ust.hk} \\
   \AND
  Ruixin Hong \\
  Department of Automation\\
  Tsinghua University\\
  Beijing, P.R.China \\
  \texttt{hrx20@mails.tsinghua.edu.cn} \\
   \And
   Yangqiu Song \\
   Department of CSE \\
   HKUST \\
   Hong Kong, P.R.China \\
   \texttt{yqsong@cse.ust.hk} \\
   \And
   Changshui Zhang \\
  Department of Automation\\
  Tsinghua University\\
  Beijing, P.R.China \\
  \texttt{zcs@mail.tsinghua.edu.cn} \\
}

\begin{document}

\maketitle

\begin{abstract}

The visual dialog task requires an AI agent to interact with humans in multi-round dialogs based on a visual environment. 
As a common linguistic phenomenon, pronouns are often used in dialogs to improve the communication efficiency. As a result, resolving pronouns (i.e., grounding pronouns to the noun phrases they refer to) is an essential step towards understanding dialogs.
In this paper, we propose VD-PCR, a novel framework to improve Visual Dialog understanding with Pronoun Coreference Resolution in both implicit and explicit ways. 
First, to implicitly help models understand pronouns, we design novel methods to perform the joint training of the pronoun coreference resolution and visual dialog tasks. 
Second, after observing that the coreference relationship of pronouns and their referents indicates the relevance between dialog rounds, we propose to explicitly prune the irrelevant history rounds in visual dialog models' input. With pruned input, the models can focus on relevant dialog history and ignore the distraction in the irrelevant one.
With the proposed implicit and explicit methods, VD-PCR achieves state-of-the-art experimental results on the VisDial dataset.
The data, code and models are available at: \url{https://github.com/HKUST- KnowComp/VD-PCR}.

\end{abstract}


\section{Introduction}

\begin{figure}[t]
  \centering
  \includegraphics[width=0.8\linewidth]{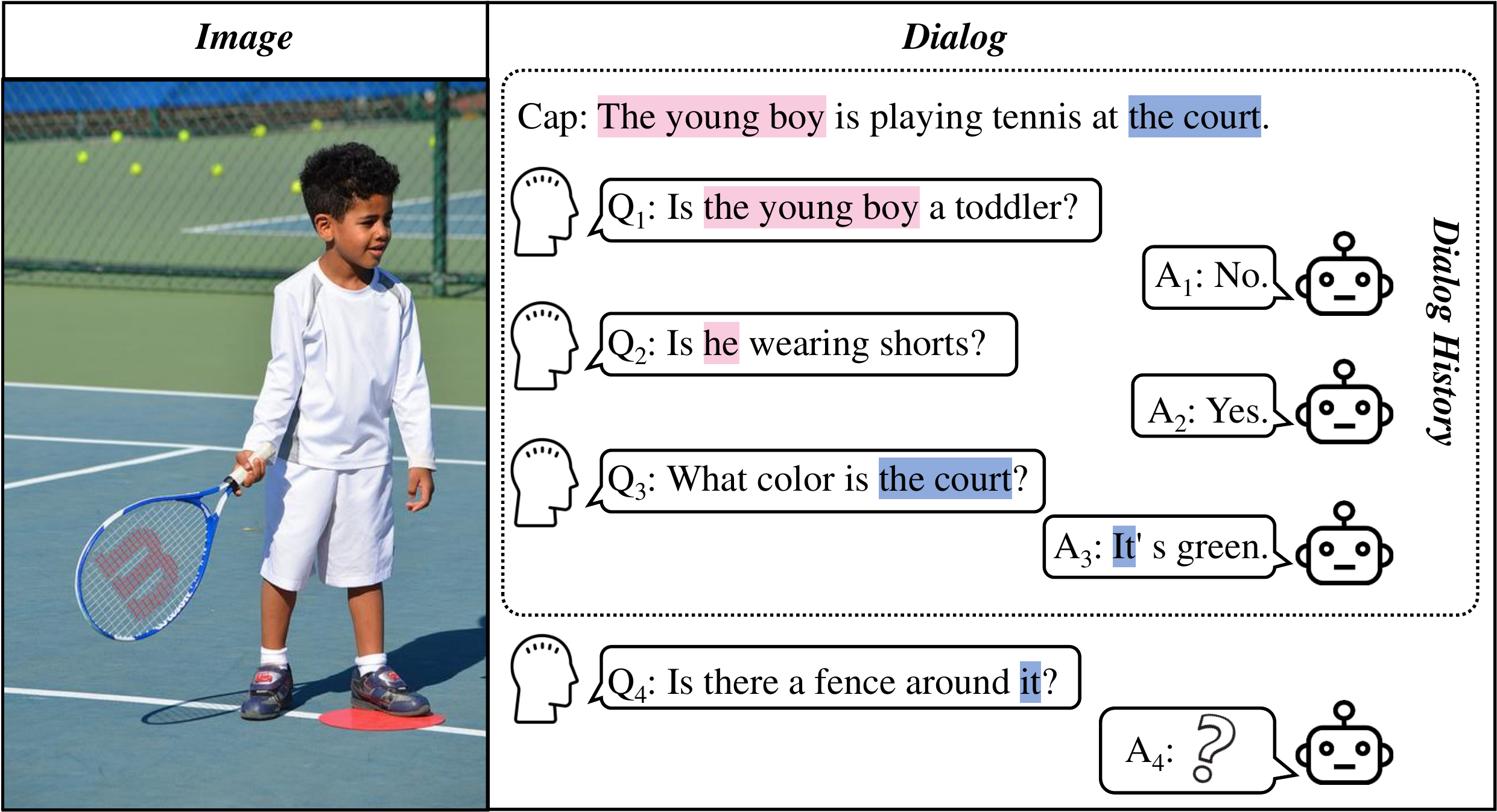}
  \caption{An example of the visual dialog task. ``Cap,'' ``Q,'' and ``A'' denote caption, question, and answer, respectively. With the image and dialog history as context, the model is required to choose the correct answer for $Q_4$. Pronouns and noun phrases that refer to the same entity are highlighted in the same color.}
  \label{fig:dialog_example1}
\end{figure}

Recently, we have witnessed substantial progress in the intersection of computer vision and natural language processing. Researchers have designed multi-modal tasks such as image captioning~\cite{DBLP:conf/acl/SoricutDSG18}, visual question answering~\cite{DBLP:conf/iccv/AntolALMBZP15}, and visual commonsense reasoning~\cite{DBLP:conf/cvpr/ZellersBFC19} to promote the mutual understanding and reasoning of vision and language modalities. 
Among all multi-modal tasks, visual dialog~\cite{DBLP:conf/cvpr/DasKGSYMPB17} is specifically proposed to mimic multi-round human-AI interaction in a multi-modal environment.
This task requires an AI agent to answer questions about an image with a dialog history of consecutive question-answer pairs.
Figure~\ref{fig:dialog_example1} shows an example from the VisDial dataset~\cite{DBLP:conf/cvpr/DasKGSYMPB17}. With a dialog history of a caption and three rounds of question answering, the AI agent is expected to answer $Q_4$ about the image.
The visual dialog models' techniques could benefit the development of chat-bot-based AI assistants and devices for visually impaired people.

In dialogs, humans frequently use pronouns to refer to the noun phrases mentioned previously to maintain a compact and consistent dialog flow.
As an essential dialog phenomenon, pronouns are also common in visual dialogs.
In VisDial, 98\% of dialogs contain at least one pronoun.
For instance, in Figure~\ref{fig:dialog_example1}, speakers use ``he'' to refer to ``the young boy'' in $Q_2$ and ``it'' for ``the court'' in $A_3$ and $Q_4$.

Although previous studies have noticed the importance of understanding pronouns in dialogs, they mostly use soft attention mechanisms to implicitly learn the relationship between words and do not explicitly ground pronouns to their referents~\cite{DBLP:conf/cvpr/NiuZZZLW19,DBLP:conf/acl/GanCKLLG19}. Recently, \citet{DBLP:conf/emnlp/YuZSSZ19} propose a dataset VisPro, which annotates pronoun coreference on a portion of VisDial dialogs. With such annotations, we can accurately resolve pronouns in VisDial.
While VisPro provides extra annotation on VisDial, incorporating pronoun coreference to improve dialog understanding is not trivial.

In this paper, we propose a novel framework to improve Visual Dialog understanding with Pronoun Coreference Resolution called VD-PCR. It contains implicit and explicit methods to incorporate pronoun coreference into visual dialog models.

First, the implicit way is to train the pronoun coreference resolution and visual dialog tasks jointly. 
Pronoun coreference resolution (PCR) is the task of grounding pronouns to their referents.
An intuitive idea for joint training is to use the same base model to extract dialogs' representations and optimize both tasks' losses simultaneously. 
Following the previous state-of-the-art visual dialog models~\cite{DBLP:conf/eccv/MurahariBPD20,DBLP:conf/emnlp/WangJLKXH20}, we use a multi-modal BERT as the base model and build feed-forward layers upon the representations for each task.
However, from the experiments, we observe that the basic joint training brings about marginal improvement for visual dialog metrics. A possible reason is that the two tasks are of different language understanding levels.
The visual dialog task requires high-level semantics understanding, while the PCR task is one of the mid-level language understanding steps.

To compatibly integrate the tasks, we propose to leverage different representations for tasks of different levels. 
In the multi-modal BERT with multiple transformer layers, the last layers' output learns complex contextual semantics, so we utilize them for the visual dialog task.
For the PCR task, we select the attention heads which capture coreference best during the pretraining among all BERT layers and use their output as representation.
In this way, the PCR supervision enhances the model's ability to understand coreference in the mid-level layers, contributing to a better high-level representation in the last layer for the visual dialog task.

Second, the explicit way to leverage pronoun coreference is to prune the dialog history in visual dialog models' input.
Although questions in VisDial are raised under dialog history as context, not all previous rounds have a direct relationship with the target questions. 
For example, the pronoun ``it'' in $Q_4$ in Figure~\ref{fig:dialog_example1} refers to ``the court'' in the caption and $Q_3$.
While the caption and $Q_3$ are directly relevant to $Q_4$, the other utterances are unnecessary for understanding $Q_4$.
Including irrelevant history as context induces harmful bias to visual dialog models.
\citet{DBLP:conf/cvpr/QiNHZ20} observe that existing models tend to favor answers that match some history words, and such word-match bias could be misleading. For instance, when all the answers in history are ``yes,'' the models would prefer ``yes'' to answer the target question even if the correct answer is ``no.''

While current works try to address this issue by directly removing all the dialog history in the input~\cite{DBLP:conf/cvpr/QiNHZ20,DBLP:conf/aaai/KimTB20},
we propose to include the minimal dialog history in the input precisely. In specific, we keep the relevant history rounds and prune the irrelevant ones so that we can have the necessary context to understand the dialog and minimize the adverse effect of dialog history.
The relevance between history rounds and the target questions can be judged by pronoun coreference.
We consider only the target question with referential pronouns and the dialog rounds with noun phrase antecedents of those pronouns as relevant.

We verify our methods with PCR experiments on the VisPro dataset and visual dialog experiments on the VisDial dataset.
Since the pronoun coreference annotation of VisPro only covers 3.75\% VisDial data, we first train a PCR model to resolve pronouns in all dialogs.
On top of the traditional text-only PCR models,
we select multi-modal BERT as the base representation model to incorporate visual information. Moreover, we leverage the vast amount of unlabelled dialogs with pseudo labels to train the PCR model on all VisDial data. With multi-modal BERT and pseudo labels, we achieve 3.4 absolute improvement on the F1 metric compared to text-only models.
With pronouns resolved by the PCR model, we apply the VD-PCR framework to the visual dialog task.
We set new state-of-the-art records with a 68.73 mean reciprocal rank (MRR) score by joint training and a 76.14 normalized discounted cumulative gain (NDCG) score by history pruning.

To summarize, our contribution is based on the following novel proposals:

\begin{itemize}
    \item To resolve pronouns in VisDial, we propose to include visual information with multi-modal BERT and train on the whole VisDial dataset with pseudo labels.
    \item We incorporate pronoun coreference into visual dialog models implicitly by the joint training of the PCR and visual dialog tasks. To compatibly integrate the two tasks, we propose to leverage different representations for tasks of different understanding levels. 
    \item We propose to explicitly prune irrelevant dialog history in visual dialog models' input. Pruning the irrelevant dialog rounds identified by pronoun coreference prevents the models from overfitting to irrelevant history context.
    \item With the help of pronoun coreference, our VD-PCR model achieves state-of-the-art results on VisDial. We conduct extensive quantitative and qualitative experiments to illustrate the effects of the proposed mechanisms.
\end{itemize}

The rest of the paper is organized as follows. Section~\ref{sec:related_works} introduces previous studies on the visual dialog and PCR tasks. Section~\ref{sec:tasks} describes the definition of the tasks and analyzes the pronoun phenomenon in VisDial. Section~\ref{sec:methods} first introduces the PCR model on VisDial and then describes VD-PCR with joint training and history pruning. Section~\ref{sec:experiments} reports experimental results and their analysis. In the end, Section~\ref{sec:conclusion} concludes the paper.

\section{Related Works}
\label{sec:related_works}
In this section, we first introduce existing works on the visual dialog and pronoun coreference resolution tasks. Then we describe previous discussion of pronoun coreference and dialog history in the visual dialog task. 

\subsection{Visual dialog}




The visual dialog task requires AI agents to answer questions with image and dialog history as contexts.
Early methods on the VisDial dataset extract representations of images, questions, and history rounds separately by CNN and LSTM. The features are then fused via attention and various mechanisms such as involving multi-step attention~\cite{DBLP:conf/acl/GanCKLLG19}, 
incorporating scene graphs~\cite{DBLP:conf/aaai/JiangYQZZ0W20}, and using variational auto-encoders~\cite{DBLP:journals/pr/PatroAN21} to perform multi-modal reasoning. 
While we focus on discriminating answers in visual dialog, the dataset is also exploited to develop AI agents to generate both questions and answers based on images and dialog history~\cite{DBLP:conf/emnlp/MurahariCBPD19,DBLP:journals/pr/ZhaoLSG21}.

Since the recent success of large-scale pretraining models such as BERT~\cite{DBLP:conf/naacl/DevlinCLT19} 
in the natural language processing community, researchers have designed multi-modal BERT models which allow both visual and textual input. They use single~\cite{DBLP:conf/aaai/LiDFGJ20,DBLP:conf/eccv/Li0LZHZWH0WCG20} or two-stream~\cite{DBLP:conf/nips/LuBPL19,DBLP:conf/emnlp/TanB19} transformer layers to compute attention between two modalities. The models are pretrained on large-scale multi-modal datasets with self-supervised objectives. Further finetuning them on specific tasks leads to new state-of-the-art records on several multi-modal challenges such as visual question answering~\cite{DBLP:conf/iccv/AntolALMBZP15,DBLP:journals/pr/FangLLQL19,DBLP:journals/prl/LiSLZF20}, image-text retrieval~\cite{DBLP:conf/eccv/LeeCHHH18,DBLP:journals/pr/LiuLGL18}, and visual commonsense reasoning~\cite{DBLP:conf/cvpr/ZellersBFC19}.
\citet{DBLP:conf/eccv/MurahariBPD20} adapt the two-stream ViLBERT~\cite{DBLP:conf/nips/LuBPL19}  to VisDial via a two-step finetuning and boost the evaluation metrics by a large margin.
\citet{DBLP:conf/emnlp/WangJLKXH20} propose a single-stream VD-BERT for visual dialog via a three-step training and achieve on-par results.


\subsection{Pronoun coreference resolution}
Coreference resolution is the task of identifying mentions referring to the same entity.
As a fundamental natural language understanding task, the accurate resolution of coreference could benefit down-stream applications such as machine translation~\cite{DBLP:conf/eacl/Guillou12}, question answering~\cite{DBLP:conf/nldb/BhattacharjeeHW20}, and dialog systems~\cite{DBLP:conf/acl/StrubeM03}.
Early methods~\cite{DBLP:conf/emnlp/RaghunathanLRCSJM10,DBLP:conf/acl/ClarkM15} leverage hand-craft rules to judge the coreference relationships between mentions, while recent works build end-to-end neural networks to solve the task based on pretrained language models such as ELMo~\cite{DBLP:conf/emnlp/LeeHLZ17,DBLP:conf/naacl/LeeHZ18}, BERT~\cite{DBLP:conf/emnlp/JoshiLZW19}, and SpanBERT~\cite{DBLP:journals/tacl/JoshiCLWZL20}.
As a subtask of the general coreference resolution, the pronoun coreference resolution (PCR) task focuses on grounding pronouns to the mentions they refer to~\cite{DBLP:conf/aaai/Ng05,DBLP:conf/naacl/ZhangSS19,DBLP:conf/acl/ZhangSSY19}. It is more challenging than the general one because pronouns have weak semantic meaning~\cite{DBLP:conf/acl/Ehrlich81}. 

For the joint training of coreference resolution and downstream applications, \citet{DBLP:conf/emnlp/QuanXWH19} convert coreference resolution into the same format of the application as a generation task. \citet{DBLP:conf/nldb/BhattacharjeeHW20} substitute the pronouns with their referents. 
Unlike previous works, our VD-PCR does not change the task's definition or texts. Instead, we propose to use different representations for tasks of different levels.

\subsection{Pronoun coreference in visual dialog}

In the LSTM-based visual dialog models, coreference between pronouns in target questions and mentions in dialog history has inspired the design of various reasoning mechanisms~\cite{DBLP:conf/emnlp/KangLZ19}. For instance, \citet{DBLP:conf/eccv/KotturMPBR18} devise neural modules to perform visual coreference resolution, which aims to ground a pronoun and the corresponding object in the image to the first mention of the object in the dialog. \citet{DBLP:conf/cvpr/NiuZZZLW19} propose eliminating the ambiguity of pronouns in questions by recursively attention on the history rounds. 
However, they model coreference by attention between words without explicitly resolving pronouns.
Moreover, none of these methods are applicable to the multi-modal BERT, which integrates all the attention computation into transformer layers.

Recently, \citet{DBLP:conf/emnlp/YuZSSZ19} annotates the pronoun coreference information on a portion of dialogs in VisDial to form a dataset VisPro. 
In this work, we leverage VisPro's the PCR labels to train a PCR model on VisDial. After explicitly grounding pronouns to their referents, we propose VD-PCR with novel mechanisms to improve BERT-based visual dialog models with pronoun coreference.

\subsection{Dialog history in visual dialog}
\label{sec:dialog_history}

As dialog history is a major difference between multi-round visual dialog and single-round VQA, the effect of dialog history in visual dialog has been heatedly discussed since the proposal of the task. 
\citet{DBLP:conf/cvpr/QiNHZ20} first analyze the harmful bias brought by the direct input of history to the answer model. They observe that models with history prefer answers with similar lengths or words with history utterances.
\citet{DBLP:conf/aaai/KimTB20} and \citet{DBLP:conf/emnlp/WangJLKXH20} report similar drop in ranking metrics when more history rounds are included.
To address the adverse effect of dialog history, they either use soft attention to distinguish the relevance of history rounds~\cite{DBLP:conf/acl/AgarwalBLKR20} or directly control the number of history rounds in input~\cite{DBLP:conf/aaai/KimTB20,DBLP:conf/emnlp/WangJLKXH20}. The former can not bypass the word-match bias favored by attention mechanisms, and the latter could lose necessary information to understand the target round.

In contrast, we propose to utilize dialog history precisely with the help of pronoun coreference. By keeping the relevant rounds and pruning the irrelevant ones judged by coreference, we provide the necessary history for dialog understanding and minimize history's negative impact simultaneously.

\section{Tasks and Datasets}
\label{sec:tasks}

In this section, we first present the formal definition of the visual dialog and PCR task. Then we give a detailed analysis of pronouns in the dialog scenario.

\subsection{The visual dialog task}
\label{sec:visdial}
In VisDial, each dialog $D$ discusses an image $I$. A dialog consists of a caption $Cap$ and several consecutive rounds of question-answering pairs $\{(Q_i, A_i) \}$. For each question $Q_t$, the task is to rank 100 candidate answers $\{A_t^{(j)}\}_{j=1}^{100}$ based on the image $I$ and the dialog history $\{ Cap, (Q_1, A_1), ..., (Q_{t-1}, A_{t-1})\}$.

The VisDial dataset provides two kinds of annotations for candidate answers, i.e., the sparse ones and the dense ones. They result in different training targets and metrics for evaluation. The sparse annotations give the ground-truth answer $A_t$ out of all candidates $\{A_t^{(j)}\}_{j=1}^{100}$. Models are trained to retrieve $A_t$ and retrieval metrics are adopted for evaluation. 
The dense annotations are the relevance scores $\{r^{(j)}\}_{j=1}^{100} \in [0, 1]$ for all candidates. Models are required to predict $\{r^{(j)}\}$ for all candidates and give a ranking by the scores. Ranking metrics are employed with dense annotations.

With different definitions and annotating processes, the two annotations are misaligned and even conflicting in some of the dialogs.
For instance, for the question ``Is it sunny?'', there are similar candidate answers such as ``yes'', ``yes, it is'', ``yes, it's sunny''. The sparse annotation may label only ``yes, it is'' as the ground truth and the other two as wrong answers, while the dense annotation gives 1.0 scores to all of them. 
In another case, the ground truth answer for the question gets a 0 score in the dense annotation.
Because of these misalignments and conflicts, current models can not optimize both metrics simultaneously, and the improvement of one metric could hurt the other one~\cite{DBLP:conf/eccv/MurahariBPD20}.
Moreover, the dense annotations only cover 0.9\% of VisDial data while the sparse ones are provided on all dialog rounds. Thus, existing methods usually split the training process into two phases. In Phase 1, the model is trained to optimize retrieval metrics with extensive sparse annotations. In Phase 2, the model from Phase 1 is further finetuned with the scarce dense annotations to maximize the ranking metrics.

\subsection{The pronoun coreference resolution task}
\label{sec:pcr_task}
Given a dialog record $D$ and a target pronoun $p$, the goal of PCR is to recognize all the noun phrases $n$ in the preceding text of $p$ that $p$ refers to and co-refers with~\cite{DBLP:conf/naacl/ZhangSS19,DBLP:conf/emnlp/YuZSSZ19}. These noun phrases are called antecedents of the pronoun. All the noun phrases previous to the pronouns are deemed as candidates of antecedents.
For example, for the pronoun ``it'' of $Q_4$ in the dialog in Figure~\ref{fig:dialog_example1}, a PCR model is expected to identify ``the court'' in the caption and ``the court'' in $Q_3$ as its antecedents among all the candidates such as ``the young boy'' in $Q_1$ and ``shorts'' in $Q_2$.
Following \cite{DBLP:conf/acl/StrubeM03,DBLP:conf/aaai/Ng05,DBLP:conf/emnlp/YuZSSZ19}, the pronouns that we discuss contain third-person personal (it, he, she, they, him, her, them) and possessive pronouns (its, his, her, their).

\subsection{Pronouns in VisDial}

The VisPro dataset~\cite{DBLP:conf/emnlp/YuZSSZ19} annotates pronoun coreference in VisDial. In each annotated dialog, it labels the noun phrase antecedents of the third-person pronouns following the definition in Section~\ref{sec:pcr_task}. 

To understand pronouns in VisDial, we first analyze their statistics.
While VisPro provides PCR annotations, it contains only 3.75\% dialogs in VisDial. Besides, a dialog must have four to ten pronouns to be selected in VisPro. Data in VisPro alone can not represent the general distribution of pronouns in VisDial.
Therefore, we leverage the predictions of our PCR model, which will be introduced in Section~\ref{sec:phase_0}, together with the annotations in VisPro to analyze the statistics of pronouns in VisDial.

We first discriminate referential and non-referential pronouns.
While referential pronouns refer to specific entities, non-referential pronouns are placeholders with no antecedent and no meaning, e.g., ``it'' in ``Is it sunny?''.
We observe that among 534,951 pronouns from 123,287 dialogs in the training set, 73.21\% are referential. 84.3\% of dialogs have at least one referential pronoun, and 57.12\% of these dialogs contain at least two pronouns referring to different entities. Among all dialog rounds, 19.35\% of questions and 10.42\% answers have at least one referential pronoun. Statistics indicate that referential pronouns are common in VisDial and that disambiguating their meaning is crucial to understanding the dialogs.

\begin{figure}[t]
  \centering
  \vspace{-0.5in}
  \includegraphics[width=0.9\linewidth]{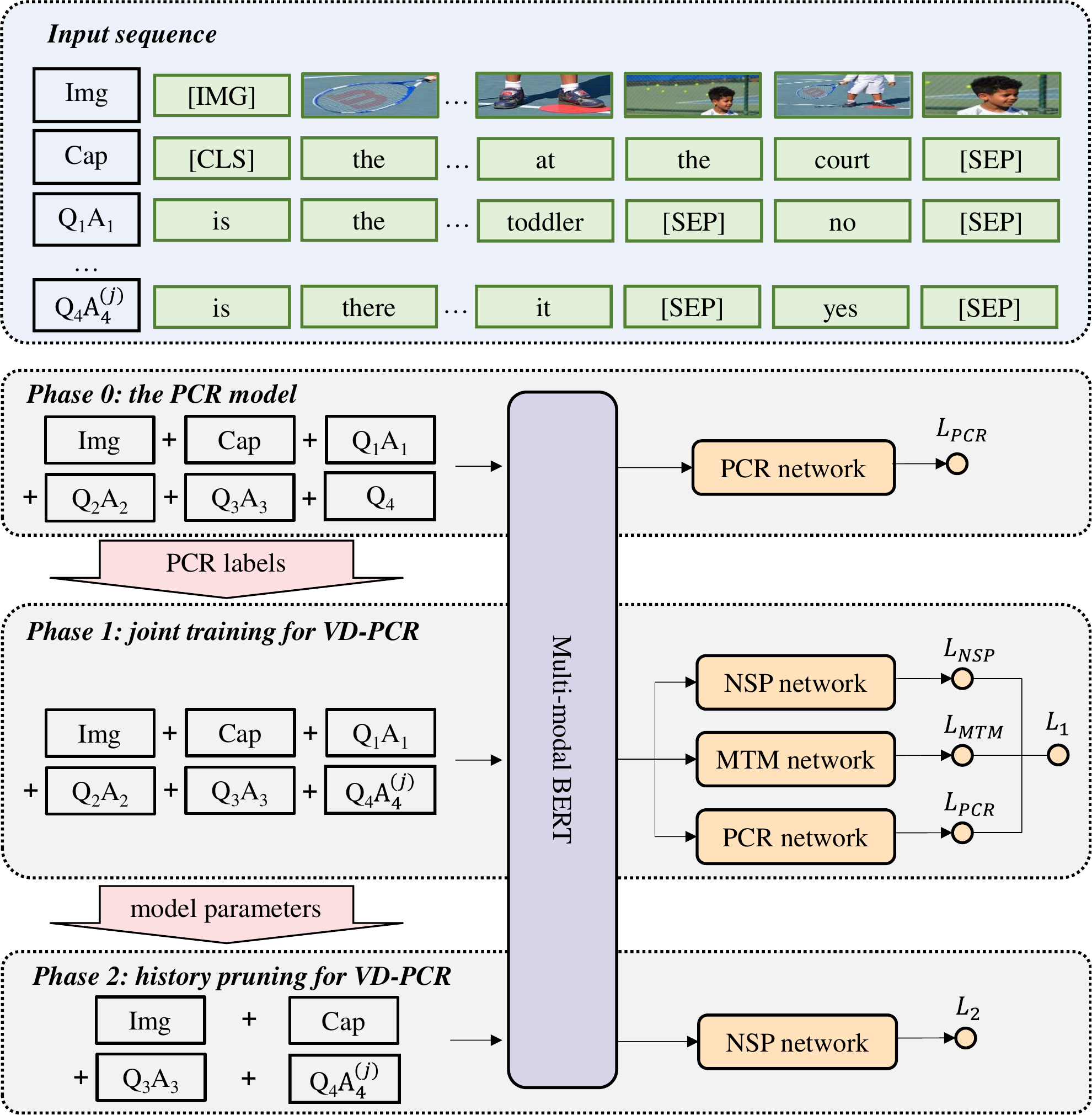}
  \caption{
  The overview of inputs and model structures for the PCR and VD-PCR models. 
  While all models use a multi-modal BERT as the base model, their input and loss functions are different.
  Take answering $Q_4$ in Figure~\ref{fig:dialog_example1} as an example.
  The PCR model for Phase 0 takes the whole dialog previous to $Q_4$ as input and optimizes $\mathcal{L}_{PCR}$.
  For the VD-PCR model in Phase 1, the input includes all dialog history. The model is jointly trained by the sum of next sentence prediction (NSP), masked token modeling (MTM), and PCR losses.
  The VD-PCR model for Phase 2 only includes relevant history rounds in input and minimizes the loss function $\mathcal{L}_2$. In this example, the relevant rounds for $Q_4$ are $\{Cap, (Q_3, A_3)\}$ judged by pronoun coreference.
  As for the relationship between phases, Phase 0 provides PCR labels for the following phases, and the Phase 2 model is trained on the parameters learned in Phase 1.
  }
  \label{fig:model_overview}
\end{figure}

\section{Methods}
\label{sec:methods}
\subsection{Overview}
\label{sec:overview}

We first train a PCR model to resolve pronouns in visual dialogs. Then we incorporate pronoun coreference to improve the visual dialog models with the VD-PCR framework. To quantitatively measure models' performance, we evaluate the PCR task on VisPro and the visual dialog task on VisDial.

As introduced in Section~\ref{sec:visdial}, the training of visual dialog models consists of Phase 1 with sparse annotations and Phase 2 with dense annotations. As PCR on VisDial prepares coreference information for the following two phases, we call it Phase 0.

The three phases all use a multi-modal BERT as the base model but vary in the input formats and feed-forward layers built on top of BERT. Figure~\ref{fig:model_overview} presents the general structures of three models. 

To accomplish the PCR task in Phase 0, we adapt PCR models to VisPro by using a multi-modal BERT as the base model to include visual information. As VisPro only covers a small fraction of VisDial, we further improve the model by using the whole VisDial as training data with pseudo labels. The prediction of the PCR model in Phase 0 are used as the PCR labels on the whole VisDial in Phase 1 and 2.

We then design different techniques to leverage coreference to suit the two training phases in the visual dialog task. In Phase 1, to retrieve the ground-truth answers, dialog history plays a vital role. The ground-truth answers usually distinguish themselves among similar candidate answers by large overlaps with the history. For instance, for $Q_4$ ``Is there a fence around it?'' in Figure~\ref{fig:dialog_example1}, ``yes'' is more likely than ``yes there is'' because the former shares the same format with $A_1$ and $A_2$ . Thus, we keep all the dialog history as input and use pronoun coreference implicitly by the joint training of PCR and visual dialog tasks. In particular, the attention heads with the strongest correspondence to coreference in the multi-modal BERT are selected as the token representations for PCR networks. 

As for Phase 2, the goal is to score the relevance of each candidate answer. For most candidates, overlaps with history have little to do with relevance. For the above example $Q_4$, consider two candidates ``no'' and ``yes, there is.'' The former has a stronger overlap with the history while the latter is more relevant. To avoid harmful bias brought by dialog history, we minimize the amount of history in the input.
Therefore, we take advantage of pronoun coreference by pruning the irrelevant dialog rounds judged by pronoun coreference.

\subsection{Phase 0: PCR model on VisDial}
\label{sec:phase_0}

\subsubsection{PCR model with multi-modal BERT}

In Phase 0, we aim to identify the coreference relationship between pronouns and noun phrases.
Unlike traditional text-only PCR tasks, the images in VisPro could help models to understand coreference in the texts better. For example, in the dialog in Figure~\ref{fig:dialog_example1}, if the model computes the attention of each textual mentions on objects in the image, it could observe a strong correlation between ``the young boy'' in the texts and the boy in the image as well as ``he'' in $Q_2$ and the boy. Thus  ``the young boy'' and ``he'' are more likely to refer to the same entity, i.e., the boy in the image.

We utilize the images by replacing the language model in the coreference model with a multi-modal BERT ViLBERT~\cite{DBLP:conf/nips/LuBPL19}. While traditional language models only take textual tokens as input, ViLBERT views image objects and dialog words as visual and textual tokens.
It incorporates visual information into the representations of language tokens via cross-modal attention, which could benefit the identification of coreference.

After ViLBERT extracts the representations $\bm{x}$ of textual tokens, we build the PCR network upon $\bm{x}$.
Following \cite{DBLP:conf/emnlp/JoshiLZW19}, for each pair of pronoun $p$ and noun phrase $n$, we calculate the coreference score $s(p, n)$ and maximize the scores of the correct $(p, n)$ pairs.
The coreference score is computed with feed-forward layers based on their representations $\bm{x}_p$ and $\bm{x}_n$:
\begin{equation}
    s(p,n) = {\rm FFN}_c(\bm{x}_{p}, \bm{x}_{n}).
\end{equation}
where ${\rm FFN}_c$ denotes a feed-forward network with two MLP layers.

For a specific pronoun $p_i$, the probability of it co-referring with a noun phrase $n_j$ is obtained with a Softmax function on the coreference score:
\begin{equation}
    P(p_i, n_j) = \frac{e^{s(p_i,n_j)}}{\sum_{n_k \in {\mathcal N}_{cd}} e^{s(p_i,n_k)}},
\end{equation}
where ${\mathcal N}_{cd}$ denotes the set of all candidate noun phrases for $p_i$.

To maximize the likelihood of correct co-referring $(p, n)$ pairs, the model minimizes the loss $\mathcal{L}_{PCR}$ as the opposite number of the log likelihood on all pronouns~\cite{DBLP:conf/emnlp/LeeHLZ17,DBLP:conf/emnlp/JoshiLZW19}:
\begin{equation}
\label{eq:crf}
    \begin{aligned}
        \mathcal{L}_{PCR} = - \sum_{i} \log \sum_{n_j \in {\mathcal N}_{cr}} P(p_i, n_j)
        = - \sum_{i} \log \frac{\sum_{n_j \in {\mathcal N}_{cr}} e^{s(p_i,n_j)}} {\sum_{n_k \in {\mathcal N}_{cd}} e^{s(p_i,n_k)}},
    \end{aligned}
\end{equation}
where ${\mathcal N}_{cr}$ is the set of correct noun phrase antecedents for $p_i$.

\subsubsection{Pseudo labels on VisDial}
\label{sec:weak_sv}

VisPro only annotates pronoun coreference on 3.75\% of VisDial dialogs. 
With a limited amount of annotated data, PCR models may fail to capture various visual dialogs' underlying structures.
Thus, we leverage the rest 96.25\% with pseudo labels:

\begin{enumerate}
    \item We train a coreference model $M_s$ on the VisPro training data ${\mathcal D}_{VP}$.
    \item We use $M_s$ to predict the coreference information for all dialogs ${\mathcal D}_{VD}$ in the training set of VisDial.
    \item We define the PCR label for the whole ${\mathcal D}_{VD}$: for those $D \in {\mathcal D}_{VP}$, we still use the VisPro annotation as ground truth; for those $D \in {\mathcal D}_{VD}$ and $D \notin {\mathcal D}_{VP}$, we take the predictions of $M_s$ as pseudo labels.
    \item We train a new coreference model $M_l$ on ${\mathcal D}_{VD}$.
\end{enumerate}

Experiments show that $M_l$ achieves better performance than $M_s$.
Finally, we use $M_l$ to resolve all pronouns in VisDial.
We combine $M_l$'s predictions and VisPro's annotations to form the PCR labels for all dialogs in VisDial as described in the above step 3. The labels are used in Phase 1 and Phase 2 of the visual dialog task.

\subsection{Phase 1: Joint training for VD-PCR}
\label{sec:phase_1}

\subsubsection{Joint training based on Multi-modal BERT}
\label{sec:phase_1_1}
In Phase 1, we help the VD-PCR model understand pronouns by jointly training the PCR and visual dialog tasks. We target at minimizing the sum of the PCR loss, the masked token modeling (MTM) loss, and the next sentence prediction (NSP) loss.
We use ViLBERT as the base model and formulate the MTM and NSP network following \cite{DBLP:conf/eccv/MurahariBPD20}, which applies ViLBERT~\cite{DBLP:conf/nips/LuBPL19} to the visual dialog task.
Then we append the PCR network on ViLBERT for the joint training.

For each candidate answer $A_t^{(j)}$ of the question $Q_t$, we concatenate the image token sequence and the dialog utterances $(Cap, Q_1, A_1, ..., Q_t, A_t^{(j)})$ together as input.
The image tokens are the CNN features of object regions in the image. A special token $\mathtt{[IMG]}$ is added at the beginning of the sequence.
For the textual input, a special token $\mathtt{[CLS]}$ is added at the start, and $\mathtt{[SEP]}$ is inserted as the separator between utterances. Examples of input sequences are displayed in Figure~\ref{fig:model_overview}.

\begin{figure}[t]
  \centering
  \includegraphics[width=0.8\linewidth]{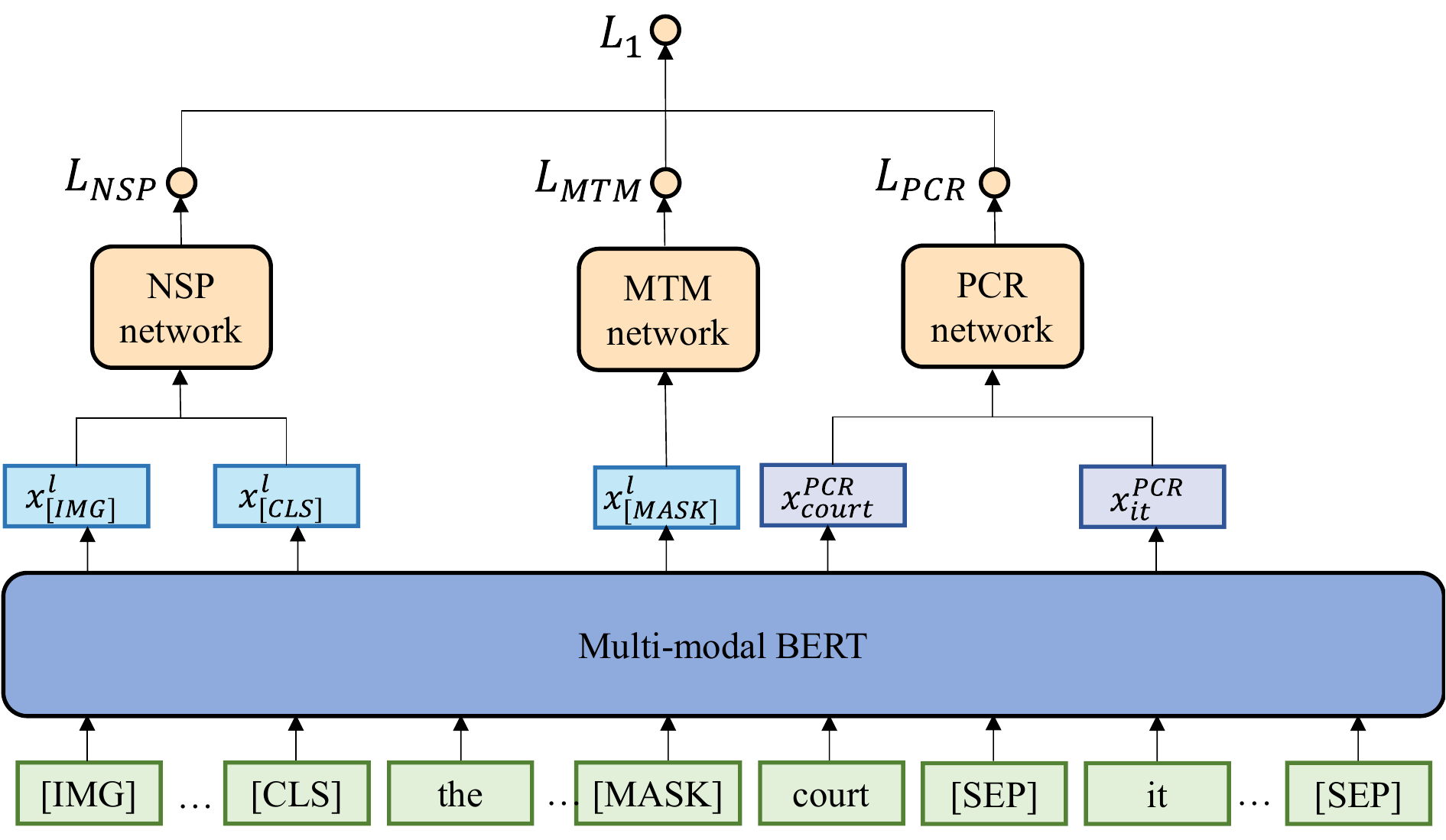}
  \caption{The structure of the joint training VD-PCR model. While the representations for the NSP and MTM networks is the output of last transformer layers ${\bm x}^l$, the PCR network's representations ${\bm x}^{PCR}$ consists of output from coreference-related attention heads as defined in Equation (\ref{eq:x_pcr}).}
  \label{fig:joint_training}
\end{figure}

The basic structure of the jointly trained model is shown in Figure~\ref{fig:joint_training}. 
For each input token $x$, the multi-modal transformers layers output its representation as ${\bm x}$.
An NSP network then performs the answer discrimination. A feed-forward neural network of one MLP layer is built on the representations of $\mathtt{[CLS]}$ and $\mathtt{[IMG]}$ to predict the probability of $A_t^{(j)}$ fitting in the dialog:
\begin{equation}
\label{eq:prob}
    [p_{A_t^{(j)}}, \bar{p}_{A_t^{(j)}}] = {\rm FFN}_d ({\bm x}^l_{\mathtt{[CLS]}} \otimes {\bm x}^l_{\mathtt{[IMG]}}),
\end{equation}
where ${\bm x}^l$ is the output of the last layer of the multi-modal BERT, $\otimes$ denotes element-wise multiplication, and $p_{A_t^{(j)}}$ and $\bar{p}_{A_t^{(j)}}$ are the scores of $A_t^{(j)}$ being the correct and wrong answer, respectively.

The label $l_{A_t^{(j)}}$ for ground-truth answers provided by the sparse annotations is 1 while the label for others is 0. Thus the NSP loss \cite{DBLP:conf/eccv/MurahariBPD20} is computed on all rounds: 
\begin{equation}
\label{eq:NSP}
    \mathcal{L}_{NSP} = - \sum_{t,j} \left[ l_{A_t^{(j)}} \log p_{A_t^{(j)}} + (1 - l_{A_t^{(j)}}) \log  \bar{p}_{A_t^{(j)}} \right] .
\end{equation}

Following \cite{DBLP:conf/eccv/MurahariBPD20}, the MTM loss is also used in the Phase 1 training.
The tokens in the input are randomly masked by the $\mathtt{[MASK]}$ token and the MTM network is trained to predict the token based on the representation of $\mathtt{[MASK]}$:
\begin{equation}
    {\bm p}_{x} = {\rm FFN}_m ({{\bm x}^l_{\mathtt{[MASK]}}}), \quad
    \mathcal{L}_{MTM} = - \sum_{x \in \mathcal{X}_m} {\bm l}_{x} \log {\bm p}_{x},
\end{equation}
where $\mathcal{X}_m$ is the set of masked tokens and ${\bm p}_{x}, {\bm l}_{x}$ are the predictions and labels of the tokens.  ${\rm FFN}_m$ contains one MLP layer.

As for the PCR task, we build the same PCR network as in Section~\ref{sec:phase_0} on ${\bm x}^l$. In total, the overall loss function for Phase 1 is
\begin{equation}
\label{eq:lambda}
    \mathcal{L}_{1} = \lambda_{NSP} \mathcal{L}_{NSP} + \lambda_{MTM} \mathcal{L}_{MTM} + \lambda_{PCR} \mathcal{L}_{PCR},
\end{equation}
where $\lambda$s are the hyper-parameters of loss coefficients.

\subsubsection{Compatible joint training by selecting coreference-related attention heads}

Ideally, including the PCR task into the visual dialog model could help the model understand pronoun coreference better and thus solve the visual dialog task better. However, in joint training experiments, we observe that directly optimizing $\mathcal{L}_{1}$ on the model described above shows only marginal improvement for visual dialog metrics.
We propose that this is due to the incompatibility of using the same BERT layer ${\bm x}^l$ as representation for tasks of different language understanding levels.
In this section, we will first introduce the basic structure of BERT, and then present how we enable compatible joint training by leveraging different BERT representations for different tasks.

\textbf{The BERT structure}

\begin{figure}[t]
  \centering
  \includegraphics[width=\linewidth]{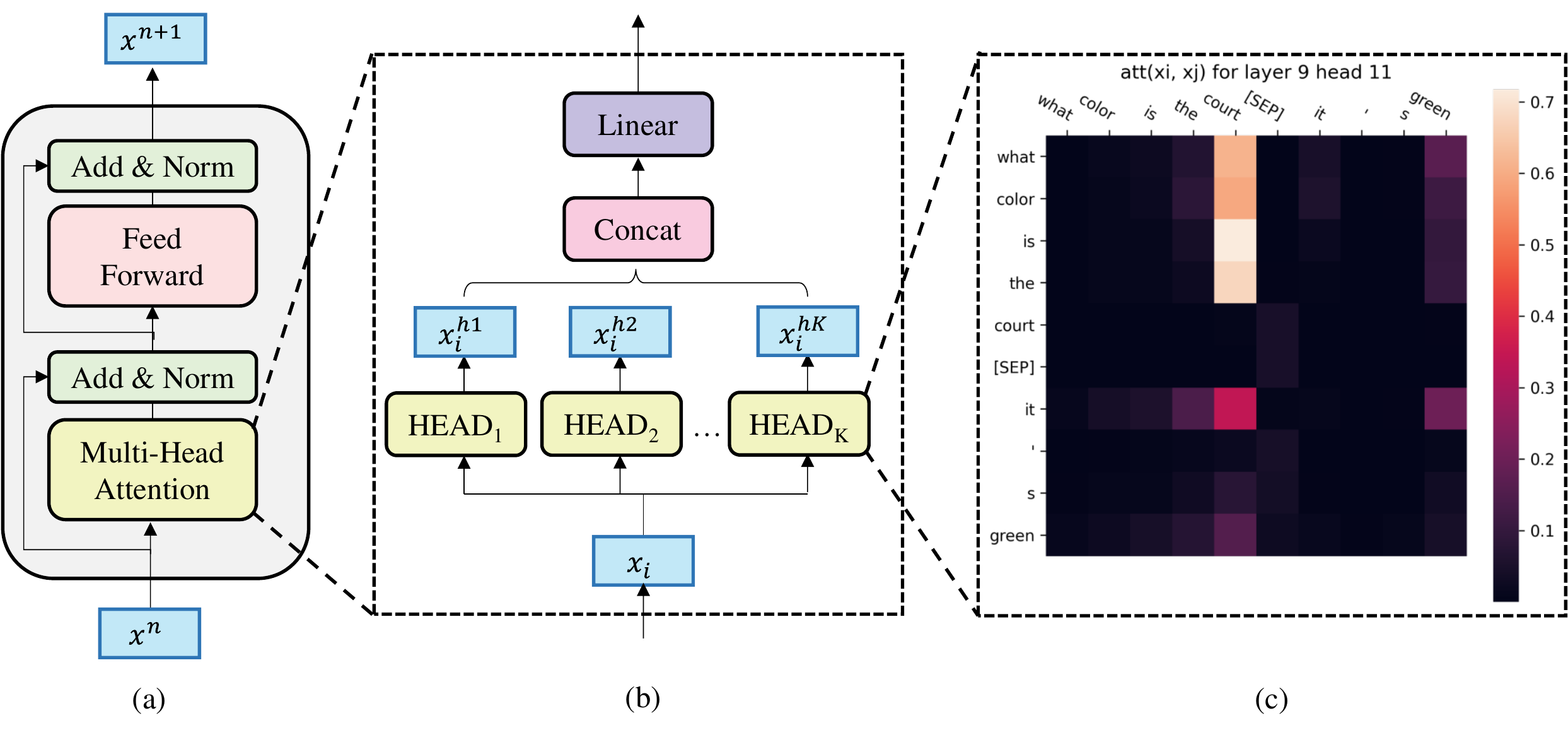}
  \caption{(a) Each transformer layer consists of a multi-head attention module and a feed-forward module. (b) The structure inside the multi-head attention module. (c) An example of the attention pattern between tokens in an attention head, which is ${\rm att}(x_i, x_j)$ defined in Equation~\ref{eq:attention}. Each row shows the attention of a token on all the tokens of the columns.}
  \label{fig:attention_calculation}
\end{figure}

The multi-modal BERT shown in Figure~\ref{fig:joint_training} consists of consecutive identical transformer layers, which output representation ${\bm x}$ for each input token $x$. For the $n^{th}$ transformer layer as presented in in Figure~\ref{fig:attention_calculation}(a), denote the input as ${\bm x}^{n}$ and the output to next layer as ${\bm x}^{n+1}$. The layer sequentially apply a multi-head attention module and two feed-forward MLP layers as a feed-forward module on the input with normalization:
\begin{equation}
    \begin{aligned}
        {\bm x}_{A}^{n} &= {\rm MultiHead}({\bm x}^{n}) \\
        \tilde{{\bm x}}^{n} &= {\rm LayerNorm}({\bm x}_{A}^{n} + {\bm x}^{n}) \\
        {\bm x}_{F}^{n} &= {\rm FFN}_t(\tilde{{\bm x}}^{n}) \\
        {\bm x}^{n+1} &= {\rm LayerNorm}({\bm x}_{F}^{n} + \tilde{{\bm x}}^{n})
    \end{aligned}
\end{equation}

Then we look into the computation inside the multi-head attention module, which is illustrated in Figure~\ref{fig:attention_calculation}(b).
It takes the token embedding ${\bm x}$ from the preceding layer as input and calculate the attention of each token on all tokens with multiple heads. Each attention head computes the attention between two tokens via scaled dot-product:
\begin{equation}
\label{eq:attention}
{\rm att}(x_i, x_j) = \frac{e^{{\bm x}_i^T {\bm x}_j}}{\sqrt{d} \sum_{x_j \in \{I, D\}} e^{{\bm x}_i^T {\bm x}_j}},
\end{equation}
where ${\bm x}_i \in \mathbb{R}^{d \times 1}$ is the embedding of $x_i$.
An example of ${\rm att}(x_i, x_j)$ of the dialog in Figure~\ref{fig:dialog_example1} is shown in Figure~\ref{fig:attention_calculation}(c), where the tokens ``what,'' ``color,'' ``is,'' ``the,'' and ``it'' have strong attention on the token ``court.''
Then the weighted embedding of $x_i$ is acquired by:
\begin{equation}
{\rm ATT}({\bm x}_i) = \sum_{x_j \in \{I, D\}} {\rm att}(x_i, x_j) {\bm x}_j.
\end{equation}
Differents heads conduct different linear transformation on the input embedding and the output of all heads are concatenated and fused by a linear layer:
\begin{align}
\label{eq:head}
    {\bm x}^{hk}_{i} &= {\rm ATT}(W^I_k {\bm x}_i),\\
    {\rm MultiHead}(x_i) &= W^O[{\bm x}^{h1}_{i}, ..., {\bm x}^{hK}_{i}].
\end{align}
where $K$ is the number of heads, $W^I_k$ and $W^O$ are parameters to learn and $[ \cdot,\cdot ]$ means concatenation.

\textbf{Compatible joint training}

BERT is made up of multiple attention heads in multiple layers. 
Researchers have examined the heads' attention patterns and discovered that different heads and layers in BERT learn to understand language from different aspects.
\citet{DBLP:journals/pnas/ManningCHKL20} find that some heads' attention patterns between tokens are much similar to coreference, while other heads' patterns may resemble other syntactic relationships.
\citet{DBLP:conf/acl/JawaharSS19} observe that the middle layers in BERT capture syntactic features while the top layers learn semantic features.

Since the visual dialog and PCR tasks are of different language understanding levels, using the same BERT representation of both tasks could induce incompatibility. To enable compatible joint training, we propose to use representations of different levels for different tasks. 
For the visual dialog task, MTM and NSP are high-level targets that require modeling comprehensive semantics, so they use the last layer of BERT as representation.
In contrast, the PCR task is a mid-level task, and coreference is learned by only some attention heads in the intermediate layers~\cite{DBLP:journals/pnas/ManningCHKL20,DBLP:conf/acl/JawaharSS19}. Thus we use the attention heads in the intermediate layers which have already captured coreference information during pretraining as representations for the PCR network.
In this way, the PCR task is exploited to strengthen the model's ability to understand coreference in the mid-level layers. It does not hinder the high-levels layer from learning complex semantics for the visual dialog task.

To select the coreference-related attention heads for the PCR task, we need to quantitatively measure the attention heads' correspondence to coreference. 
While \citet{DBLP:journals/pnas/ManningCHKL20} measure attention heads' correspondence to coreference by attention between single tokens, we propose to view a dialog as a whole and measure the correspondence by attention between groups of co-referring mentions.
For the attention heads that focus on coreference, the attention between co-referring mentions should be larger than that between mentions without a coreference relationship.
Based on the annotation of VisPro, we group the noun phrases and pronouns that refer to the same entity into coreference clusters. For example, in the dialog in Figure~\ref{fig:dialog_example1}, we can group the mentions into two clusters: (the young boy, the young boy, he) and (the court, the court, it, it).

We then gather the attention of mentions within and across clusters.
As a mention may contain multiple tokens $m_i = (x_{i1}, ..., x_{iN})$, we first average the attention of tokens to obtain the attention between mentions:
\begin{equation}
    {\rm att}(m_i, m_j) = \frac{1}{N_i N_j} \sum_{x_{in} \in m_i} \sum_{x_{jn} \in m_j} {\rm att}(x_i, x_j),
\end{equation}
where ${\rm att}(x_i, x_j)$ is defined in Equation (\ref{eq:attention}) and $N_i, N_j$ indicate the number of tokens in $m_i$ and $m_j$, respectively.

Then for a dialog $D$ with $N_C$ clusters, we have the attention of mentions within one cluster $C_l$ and across clusters $(C_l, C_{l'})$ as:
\begin{equation}
\begin{aligned}
    a_w &= \frac{1}{N_C} \sum_{C_l \in D} \ \sum_{m_i \in C_l, m_j \in C_l, m_i \neq m_j}  {\rm att}(m_i, m_j), \\
    a_a &= \frac{2}{N_C (N_C - 1)} \sum_{C_l \in D, C_{l'} \in D, C_l \neq C_{l'}} \  \sum_{m_i \in C_l, m_j \in C_{l'}}  {\rm att}(m_i, m_j).
\end{aligned}
\end{equation}

\begin{figure}[t]
  \centering
  \includegraphics[width=\linewidth]{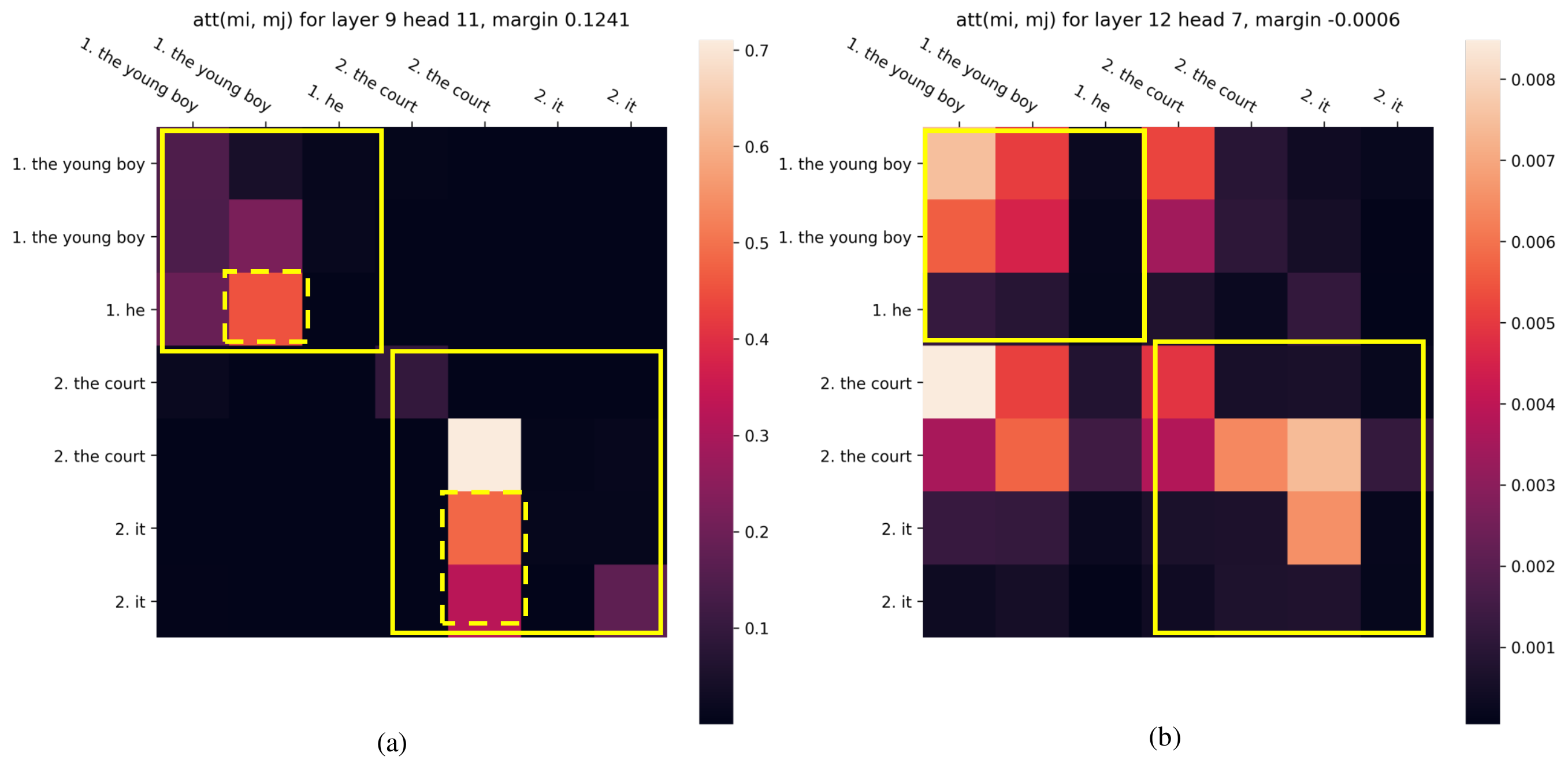}
  \caption{The attention maps ${\rm att}(m_i, m_j)$ of two heads with large and small values of $margin = a_w - a_a$ in ViLBERT on the coreference clusters of Figure~\ref{fig:dialog_example1}. Each row shows the attention of a mention on all the mentions of the columns. The two clusters are illustrated with solid-line boxes. The strong attention of pronouns to their referents are illustrated with dashed boxes.
    }
  \label{fig:attention_heads}
\end{figure}

Figure~\ref{fig:attention_heads} presents examples of ${\rm att}(m_i, m_j)$ in a coreference-related head and an uncorrelated one. The cells inside solid-line boxes are attentions within coreference clusters that sum up to $a_w$, while the attentions in the rest cells sum up to $a_a$.
Figure~\ref{fig:attention_heads}(a) shows an attention pattern similar to coreference. For instance, the dashed boxes illustrate that the attention of ``he'' on ``the young boy'' and ``it'' on ``the court'' is much stronger than the others. For this head, $a_w$ should be much larger than $a_a$.
Figure~\ref{fig:attention_heads}(b) shows another attention pattern which is uncorrelated with coreference. For this head, $a_w$ should be close to $a_a$.

Therefore, for the attention heads that capture coreference structure, $a_w$ should be much larger than $a_a$.
We average the value of $a_w - a_a$ for the dialogs in the training set of VisPro with more than one coreference clusters. 
Finally, we pick the attention heads with $a_w - a_a > a_{thres}$ as the representation for the PCR network:
\begin{equation}
\label{eq:x_pcr}
    {\bm x}^{PCR} = [{\bm x}^{h1}, ..., {\bm x}^{hK}], \  {\rm where} \  a_w^{hk} - a_a^{hk} > a_{thres},
\end{equation}
and ${\bm x}^{hk}$ is defined in Equation (\ref{eq:head}).

\subsection{Phase 2: History pruning for VD-PCR}
\label{sec:phase_2}
In Phase 2, the VD-PCR model predicts the relevance scores $\{r^{(j)}\}_{j=1}^{100}$ for all candidate answers. The output of the NSP network, i.e., the  probabilities $p_{A^{(j)}}$ in Equation (\ref{eq:prob}), are normalized and used to fit the normalized relevance scores by cross-entropy loss:
\begin{align}
    \hat{p}_{A^{(j)}} = \frac{e^{p_{A^{(j)}}}}{\sum_j e^{p_{A^{(j)}}}}&, \quad
    \hat{r}^{(j)} = \frac{e^{r^{(j)}}}{\sum_j e^{r^{(j)}}}, \\
    \mathcal{L}_{2} = - \sum_{j} &\hat{r}^{(j)} \log \hat{p}_{A^{(j)}}. 
\end{align}

As Phase 2 models have to adapt to $\mathcal{L}_{2}$ with 0.9\% VisDial data, previous works do not include MTM or NSP loss as part of $\mathcal{L}_{2}$~\cite{DBLP:conf/eccv/MurahariBPD20,DBLP:conf/emnlp/WangJLKXH20}. We do not modify the loss function either because such modification might slow down the adaptation.
Instead of adding PCR loss to $\mathcal{L}_{2}$, we make use of pronoun coreference explicitly by history pruning.

\subsubsection{History pruning mechanism}

\label{sec:pruning}

\begin{figure}[t]
  \centering
  \includegraphics[width=0.7\linewidth]{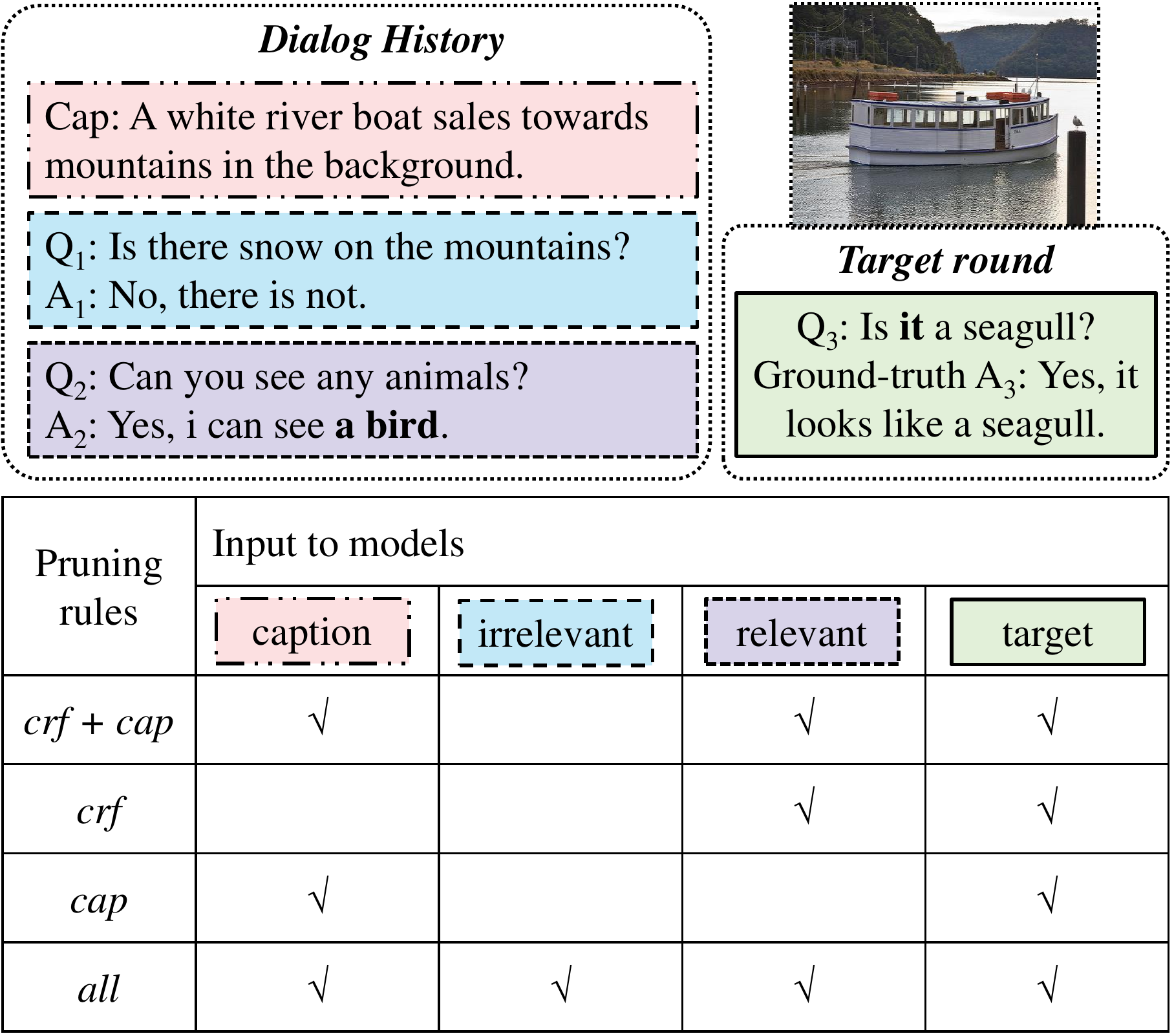}
  \caption{Illustration of different rules to prune history. The pronoun ``it'' in $Q_3$ refers to ``a bird'' in $A_2$. For the target round $(Q_3,A_3)$ in the dialog, the caption, the relevant $(Q_2, A_2)$ and irrelevant round $(Q_1, A_1)$ identified by pronoun coreference are presented in different boxes and colors. The pruned inputs defined by different rules are shown in the table.}
  \label{fig:pruning_rule}
\end{figure}

In predicting each candidate's relevance, the dialog history could bring about harmful bias.
Ideally, we can keep the relevant dialog history as input only and prune all the other rounds to alleviate the bias.
However, we do not have annotations for the dependency between questions and history rounds. We propose to judge the relevance between dialog rounds with pronoun coreference. 

To understand questions with referential pronouns, we need to ground the pronouns to noun phrases in previous rounds. 
The dialog rounds with the noun phrase antecedents are necessary for understanding the questions, while the dialog rounds without are unnecessary.
Thus we consider the former as relevant and the latter as irrelevant. For instance, in Figure~\ref{fig:pruning_rule}, the pronoun ``it'' in $Q_3$ refers to ``a bird'' in $A_2$. $(Q_2, A_2)$ is relevant to $Q_3$, and the caption and $(Q_1, A_1)$ are irrelevant. With relevance judged by pronoun coreference, we prune the original input for $Q_3$ from $(Cap, Q_1, A_1, Q_2, A_2, Q_3, A_3^{(j)})$ to $(Q_2, A_2, Q_3, A_3^{(j)})$.

In contrast, the target questions without referential pronouns can be understood without dialog history.
Take $Q_2$ in Figure~\ref{fig:pruning_rule} as an example. We can answer it by simply looking at the image.
Since all the dialog history is deemed irrelevant, these questions' input is simply the target round.

With the history pruning mechanism, we keep the minimal and necessary rounds for dialog understanding and prune the unnecessary rounds to minimize history's negative impacts.

\subsubsection{Strategies for history pruning}
\label{sec:pruning_four}

While we use the PCR model's prediction to indicate history's relevance, the predictions could be wrong. For instance, if the model fails to recognize the antecedents of the pronoun ``it'' in $Q_4$ of Figure~\ref{fig:dialog_example1}, the input for $Q_4$ will be ``Is there a fence around it?'', which is hard to understand. To alleviate incorrect PCR problem, we keep the captions as context no matter whether the PCR model predicts relevant noun phrases in them. This is because we observe that 83.16\% of pronouns in questions have noun phrase antecedents in the captions. 
When the PCR model makes mistakes, there is still a chance to include the noun phrase antecedents with captions in the input.
In the previous example, when the caption is included by default, we could still understand $Q_4$ because the caption contains ``the court.''

Besides keeping the captions and relevant dialogs round indicated by coreference, we can also define other rules to control the models' input.
In total, we experiment with four rules as illustrated in Figure~\ref{fig:pruning_rule}. 
Models trained by different rules could understand the questions from various aspects. 
The \textit{all} rule keeps all dialog history to include all information for understanding the target question, but the unnecessary rounds could induce harmful bias.
The \textit{crf} and \textit{cap} rules include only predicted co-referring rounds or captions, respectively. While the unnecessary history is substantially reduced, the necessary information might also be pruned when the pronouns in the target round refer to noun phrases out of the predicted rounds or the captions.
Thus, we define the \textit{crf + cap} rule to take the caption, relevant rounds, and the target round as input. As analyzed above, the captions and the relevant rounds could largely cover the necessary information to comprehend pronouns in target questions.
In the end, we ensemble these models to combine their advantages. The ensemble is done by averaging the normalized scores.

\section{Experiments}
\label{sec:experiments}
In this section, we first describe the experiment setup, including the datasets, evaluation metrics, baseline methods, and implementation details. Then we present the experimental results and detailed analysis of the PCR task and our VD-PCR framework of both phases in the visual dialog task.

\subsection{Experiment settings}

\subsubsection{Dataset and evaluation metrics}

We perform PCR experiments on the VisPro dataset and visual dialog experiments on the VisDial v1.0 dataset. VisDial v1.0 contains 123,287/2,064/ 8,000 dialogs of train/validation/test set\footnote{The labels for the test set are not publicly available and the evaluation on test set is done on a online server. While we report evaluation scores on the test set, we conduct the analysis that requires labels on the validation set.}, and VisPro contains 4,617/76/307 dialogs of train/validation/test set from the corresponding set of VisDial, respectively. While the sparse annotations of VisDial v1.0 cover all rounds of all dialogs, the dense annotations are provided on only one round of 2,000 dialogs in the training set and all dialogs in the validation and test set.

For evaluation on VisPro, as it annotates the coreference between pronouns $p$ and given noun phrases $n$, we view the coreference prediction of each $(p, n)$ pair as a classification problem and evaluate models with the precision, recall, and F1 score on all $(p, n)$ pairs.
For the evaluation of VisDial, the retrieval metrics contain Recall@1, 5, 10, mean reciprocal rank (MRR), and mean rank of the correct answer (MR). The ranking metric is the normalized discounted cumulative gain (NDCG).
During evaluation on the validation and test sets of the visual dialog task, we leverage the pronoun coreference information from PCR models' predictions instead of VisPro's labels.

\subsubsection{Baseline models}

For the PCR task on VisPro, we use the end-to-end coreference resolution model based on ELMo~\cite{DBLP:conf/naacl/LeeHZ18}, BERT~\cite{DBLP:conf/emnlp/JoshiLZW19}, and SpanBERT~\cite{DBLP:journals/tacl/JoshiCLWZL20} as baselines\footnote{We do not compare with VisCoref~\cite{DBLP:conf/emnlp/YuZSSZ19} because the task setting in \cite{DBLP:conf/emnlp/YuZSSZ19} excludes captions from dialogs, which is different from the visual dialog setting.}.

In the visual dialog task, we distinguish the different components in the VD-PCR with different names. We denote the Multi-modal BERT that our models base on as ``MB.'' While we mainly use ViLBERT to carry out experiments, it can be substituted by other multi-modal BERT.
In Phase 1, the Joint model with the PCR network built on MB's last layer is denoted as ``MB-J'' and the model with Coreference-related attention heads as ``MB-JC.''
For Phase 2, we apply History Pruning on the model in Phase 1 and represent it by ``HP.''

\subsubsection{Implementation details}

For the PCR model in Phase 0,
we initialize the weights of ViLBERT with \cite{DBLP:conf/eccv/MurahariBPD20} and train the model on VisPro for ten epochs\footnote{All models in our experiments converge within ten epochs.} to obtain $M_s$. After acquiring pseudo labels on VisDial, we train the model for another ten epochs to get $M_l$.

In the joint training in Phase 1, the losses coefficients defined in Equation (\ref{eq:lambda}) are $\lambda_{NSP}=1,\lambda_{MTM}=1, \lambda_{PCR} = 0.1$. 
We follow \cite{DBLP:conf/eccv/MurahariBPD20} to set the same coefficient for NSP and MTM
and search for the best coefficient for PCR from 0 to 1.
The threshold for selecting coreference-related attention heads is set to $a_{thres} = 0.1$ and there are five heads above $a_{thres}$ to be selected as the PCR network's representation.
The model is trained on the joint loss for ten epochs.


In Phase 2, we follow \cite{DBLP:conf/acl/AgarwalBLKR20} to correct the relevance scores of the ground-truth answers in sparse annotations to 1.0 during training, which could improve the retrieval metrics.

\begin{table}[t]
  \centering
  \caption{Results of PCR task on VisPro test set. We evaluate the models with precision (P), recall (R), and F1 score. Best scores are in \textbf{bold} font. ``MB'' indicates models based on multi-modal BERT. For our VD-PCR models, ``MB-J'' and ``MB-JC'' denote our jointly trained models without and with coreference-related attention head selection. Models with ``+ pseudo labels'' are trained on the whole VisDial by including the VisDial data without VisPro annotations with pseudo labels. The result with $\ddag$ is significantly higher than results with $\dag$ for p-value $<$ 0.01 (t-test).}
  \label{tab:pcr}
   \vspace{0.1in}
    \begin{tabular}{c|c|c|ccc}
    \toprule
    \multicolumn{1}{c}{Phase} & \multicolumn{2}{c}{Model} & P     & R     & F1 \\
    \midrule
    \multirow{5}[4]{*}{0} & \multicolumn{2}{c|}{ELMo~\cite{DBLP:conf/naacl/LeeHZ18}} & 91.22 & 76.52 & 83.22$\dag$ \\
          & \multicolumn{2}{c|}{BERT~\cite{DBLP:conf/emnlp/JoshiLZW19}} & 89.47 & 86.34 & 87.88$\dag$ \\
          & \multicolumn{2}{c|}{SpanBERT~\cite{DBLP:journals/tacl/JoshiCLWZL20}} & 86.68 & 87.80 & 87.24$\dag$ \\
\cmidrule{2-6}          & \multicolumn{2}{c|}{MB} & 92.81 & 86.18 & 89.37$\dag$ \\
          & \multicolumn{2}{c|}{MB + pseudo labels} & 92.67 & \textbf{88.66} & \textbf{90.62}$\ddag$ \\
    \midrule
    \multirow{2}[2]{*}{1} & \multirow{2}[2]{*}{VD-PCR} & MB-J + pseudo labels & 93.01 & 88.24 & 90.56 \\
          &       & MB-JC + pseudo labels & \textbf{94.12} & 86.06 & 89.91 \\
    \bottomrule
    \end{tabular}%
  \vspace{-0.1in}
\end{table}%

\subsection{Results of the PCR task}

Table~\ref{tab:pcr} presents the results of the PCR task on the test set of VisPro.
For the models trained with VisPro only, we can observe that the model based on MB outperforms the baseline models. It illustrates the positive impact of visual information in resolving pronouns in visual dialogs. 
By further leveraging the whole VisDial by including the VisDial data without VisPro annotations with pseudo labels, the ``MB + pseudo labels'' models boosts the F1 score significantly, which indicates the effectiveness of pseudo labels to take advantage of vast unlabelled data.
We also report the PCR results of models jointly trained for the PCR and visual dialog tasks in Phase 1.
The MB-J model, which builds the PCR network on MB's last layer, achieves on-par F1 score with the PCR-only model MB. The MB-JC model, which builds the PCR network on coreference-related attention heads, gets a lower F1 score than MB. This is because the aim of the PCR training in MB-JC is to improve dialog understanding.  Although it performs worse in the PCR task, it gets higher scores in the visual dialog task.

\begin{table}[ht]
  \centering
  \caption{Results on the test-standard phase of VisDial v1.0. $\uparrow$ indicates higher the better, and $\downarrow$ indicates lower the better. $*$ denotes models with ensemble. Best results are in \textbf{bold} font. For our VD-PCR models, ``MB-J'' and ``MB-JC'' denote the jointly trained model without and with coreference-related attention head selection. ``HP'' denotes history pruning. Results with $\ddag$ are significantly better than their baselines of the same phase with $\dag$ for p-value $<$ 0.01 (t-test).}
  \vspace{0.1in}
    \begin{tabular}{c|c|c|cccccc}
    \toprule
    \multicolumn{1}{c}{Phase} & \multicolumn{2}{c}{Model} & NDCG$\uparrow$ & MRR$\uparrow$ & R@1$\uparrow$ & R@5$\uparrow$ & R@10$\uparrow$ & MR$\downarrow$ \\
    \midrule
    \multirow{9}[2]{*}{1} & \multicolumn{2}{c|}{CorefNMN~\cite{DBLP:conf/eccv/KotturMPBR18}} & 54.70  & 61.50  & 47.55  & 78.10  & 88.80  & 4.40  \\
          & \multicolumn{2}{c|}{RvA~\cite{DBLP:conf/cvpr/NiuZZZLW19}} & 55.59  & 63.03  & 49.03  & 80.40  & 89.83  & 4.18  \\
          & \multicolumn{2}{c|}{PDUN~\cite{DBLP:journals/pr/PatroAN21}} & -     & 62.20  & 47.00  & 82.40  & 92.30  & 3.81  \\
          & \multicolumn{2}{c|}{DAN$^*$~\cite{DBLP:conf/emnlp/KangLZ19}} & 59.36  & 64.92  & 51.28  & 81.60  & 90.88  & 3.92  \\
          & \multicolumn{2}{c|}{ReDAN$^*$~\cite{DBLP:conf/acl/GanCKLLG19}} & 64.47  & 53.73  & 42.45  & 64.68  & 75.68  & 6.64  \\
          & \multicolumn{2}{c|}{DualVD~\cite{DBLP:conf/aaai/JiangYQZZ0W20}} & 56.32  & 63.23  & 49.25  & 80.23  & 89.70  & 4.11  \\
          & \multicolumn{2}{c|}{Modality-Balanced$^*$~\cite{DBLP:conf/aaai/KimTB20}} & 59.49  & 64.40  & 50.90  & 81.18  & 90.40  & 3.99  \\
          & \multicolumn{2}{c|}{ViLBERT~\cite{DBLP:conf/eccv/MurahariBPD20}} & 63.87  & 67.50$\dag$  & 53.85$\dag$  & 84.68$\dag$  & 93.25$\dag$  & 3.32$\dag$  \\
          & \multicolumn{2}{c|}{VD-BERT~\cite{DBLP:conf/emnlp/WangJLKXH20}} & 59.96  & 65.44  & 51.63  & 82.23  & 90.68  & 3.90  \\
    \midrule
    \multirow{6}[2]{*}{2} & \multicolumn{2}{c|}{MCA~\cite{DBLP:conf/acl/AgarwalBLKR20}} & 72.47  & 37.68  & 20.67  & 56.67  & 72.12  & 8.89  \\
          & \multicolumn{2}{c|}{LF+P1+P2~\cite{DBLP:conf/cvpr/QiNHZ20}} & 71.60  & 48.58  & 35.98  & 62.08  & 77.23  & 7.48  \\
          & \multicolumn{2}{c|}{P1+P2$^*$~\cite{DBLP:conf/cvpr/QiNHZ20}} & 74.91  & 49.13  & 36.68  & 62.96  & 78.55  & 7.03  \\
          & \multicolumn{2}{c|}{ViLBERT~\cite{DBLP:conf/eccv/MurahariBPD20}} & 74.47$\dag$  & 50.74$\dag$  & 37.95$\dag$  & 64.13$\dag$  & 80.00$\dag$  & 6.28$\dag$  \\
          & \multicolumn{2}{c|}{VD-BERT~\cite{DBLP:conf/emnlp/WangJLKXH20}} & 74.54  & 46.72  & 33.15  & 61.58  & 77.15  & 7.18  \\
          & \multicolumn{2}{c|}{VD-BERT$^*$~\cite{DBLP:conf/emnlp/WangJLKXH20}} & 75.35  & 51.17  & 38.90  & 62.82  & 77.98  & 6.69  \\
    \midrule \midrule
    \multirow{2}[2]{*}{1} & \multirow{6}[4]{*}{VD-PCR} & MB-J  & 62.67  & 68.04  & 54.50  & 84.88  & 93.50  & 3.27  \\
          &       & MB-JC & 63.55  & \textbf{68.73}$\ddag$ & \textbf{55.45}$\ddag$ & \textbf{85.38}$\ddag$ & \textbf{93.53}$\ddag$ & \textbf{3.21}$\ddag$ \\
\cmidrule{1-1}\cmidrule{3-9}    \multirow{4}[2]{*}{2} &       & MB-J-HP & 74.96  & 55.89  & 44.57  & 68.13  & 81.43  & 6.10  \\
          &       & MB-JC-HP & 75.30  & 56.17  & 45.32  & 68.05  & 80.98  & 6.15  \\
          &       & MB-J-HP$^*$ & 75.51 & 55.71 & 44.25 & 68.10 & 82.30 & 5.95 \\
          &       & MB-JC-HP$^*$ & \textbf{76.14}$\ddag$ & 56.05$\ddag$  & 44.75$\ddag$  & 68.40$\ddag$  & 82.75$\ddag$  & 5.72$\ddag$ \\
    \bottomrule
    \end{tabular}%
  \label{tab:vd-main-1}%
  \vspace{-0.1in}
\end{table}%

\subsection{Results of the visual dialog task}

Results of the visual dialog task on the test-standard set of VisDial v1.0 are shown in Table~\ref{tab:vd-main-1}.
For Phase 1 models, the target is to maximize retrieval metrics.
Compared to the ViLBERT baseline, our joint training model MB-JC reports substantial improvement and sets a new state-of-the-art MRR record of 68.73. 
In Phase 2, the models optimize the ranking metric NDCG.
Our MB-JC-HP model achieves the best NDCG of 75.30 among all single models, and the ensembled version achieves the state-of-the-art NDCG of 76.14 among all published results.
The results illustrates the effectiveness of the proposed VD-PCR.
We give a detailed analysis of the two phases in the following sections.

\subsection{Analysis of joint training in Phase 1}

Table~\ref{tab:vd-main-1} shows that while the baseline model ViLBERT obtains 67.50 MRR in Phase 1, the basic joint training model MB-J gets a higher MRR of 68.04 but a much lower NDCG. The slight improvement in MRR and the decrease in NDCG indicate the effectiveness of incorporating pronoun coreference implicitly via joint training and the incompatibility between the tasks.
In contrast, the MB-JC model with coreference-related heads for the PCR network achieves a slightly worse NDCG and a significantly higher MRR at 68.73. 
Moreover, in Phase 2, the MB-J-HP model improves NDCG from 74.47 of ViLBERT baseline to 74.96, and the MB-JC-HP model achieves the best NDCG of 75.30 among single models.
For ensembled models, MB-JC-HP$^*$ obtains better scores than MB-J-HP$^*$ in both ranking and retrieval metrics.
The improvement from MB-J to MB-JC in both phases further indicates the impact of using different representations for tasks of different understanding levels.

\subsubsection{Results on dialogs with different number of coreference clusters}
\label{sec:mrr_clusters}

\begin{table}[t]
  \centering
  \caption{The difference of MRR scores between ViLBERT and VD-PCR with joint training and coreference-related heads (MB-JC) on the VisDial v1.0 validation set. Dialogs are divided by the number of coreference clusters in them.}
  \vspace{0.1in}
    \begin{tabular}{c|c|p{1.5cm}<{\centering}|p{1.5cm}<{\centering}|p{1.5cm}<{\centering}}
    \toprule
    \multicolumn{1}{c}{\multirow{2}[4]{*}{\#clusters}} & \multicolumn{1}{c}{\multirow{2}[4]{*}{\#dialogues}} & \multicolumn{2}{c}{MRR} & \multirow{2}[4]{*}{$\Delta$MRR} \\
\cmidrule{3-4}    \multicolumn{1}{c}{} & \multicolumn{1}{c}{} & \multicolumn{1}{c}{ViLBERT} & \multicolumn{1}{c}{MB-JC} &  \\
    \midrule
    0     & 308   & 73.30  & 73.65  & 0.35  \\
    1     & 806   & 70.01  & 70.86  & 0.85  \\
    2     & 592   & 67.67  & 68.37  & 0.70  \\
    3     & 260   & 66.59  & 67.13  & 0.54  \\
    4-6   & 94    & 63.56  & 65.54  & 1.98  \\
    \midrule
    all   & 2060  & 69.10  & 69.85  & 0.75  \\
    \bottomrule
    \end{tabular}%
  \label{tab:mrr_diff_dialog}%
  \vspace{-0.1in}
\end{table}%

To analyze the influence of the joint training, we compare the MRR scores of ViLBERT and our model on dialogs with different numbers of coreference clusters\footnote{We only compare 2,060 dialogs out of the 2,064 dialogs in the validation set of VisDial v1.0 because ViLBERT\cite{DBLP:conf/eccv/MurahariBPD20} removes four dialogs during pre-processing for they are too long to fit in the input.}. As shown in Table~\ref{tab:mrr_diff_dialog}, for both models, with more coreference clusters in dialogs, MRR gets lower. It indicates that dialogs with more complex coreference phenomenon are harder for models to understand. For dialogs without coreference, the jointly trained model gets a slightly higher MRR score with the ViLBERT. For dialogs with coreference clusters, the gaps between the models are much larger. In particular, for dialogs with more than four clusters, the joint training increases the MRR score by a large margin of 1.98. It suggests that the improvement of joint training results from a better comprehension of coreference in dialogs.

\subsubsection{Results of different loss coefficients}

\begin{figure}[t]
  \centering
  \includegraphics[width=0.7\linewidth]{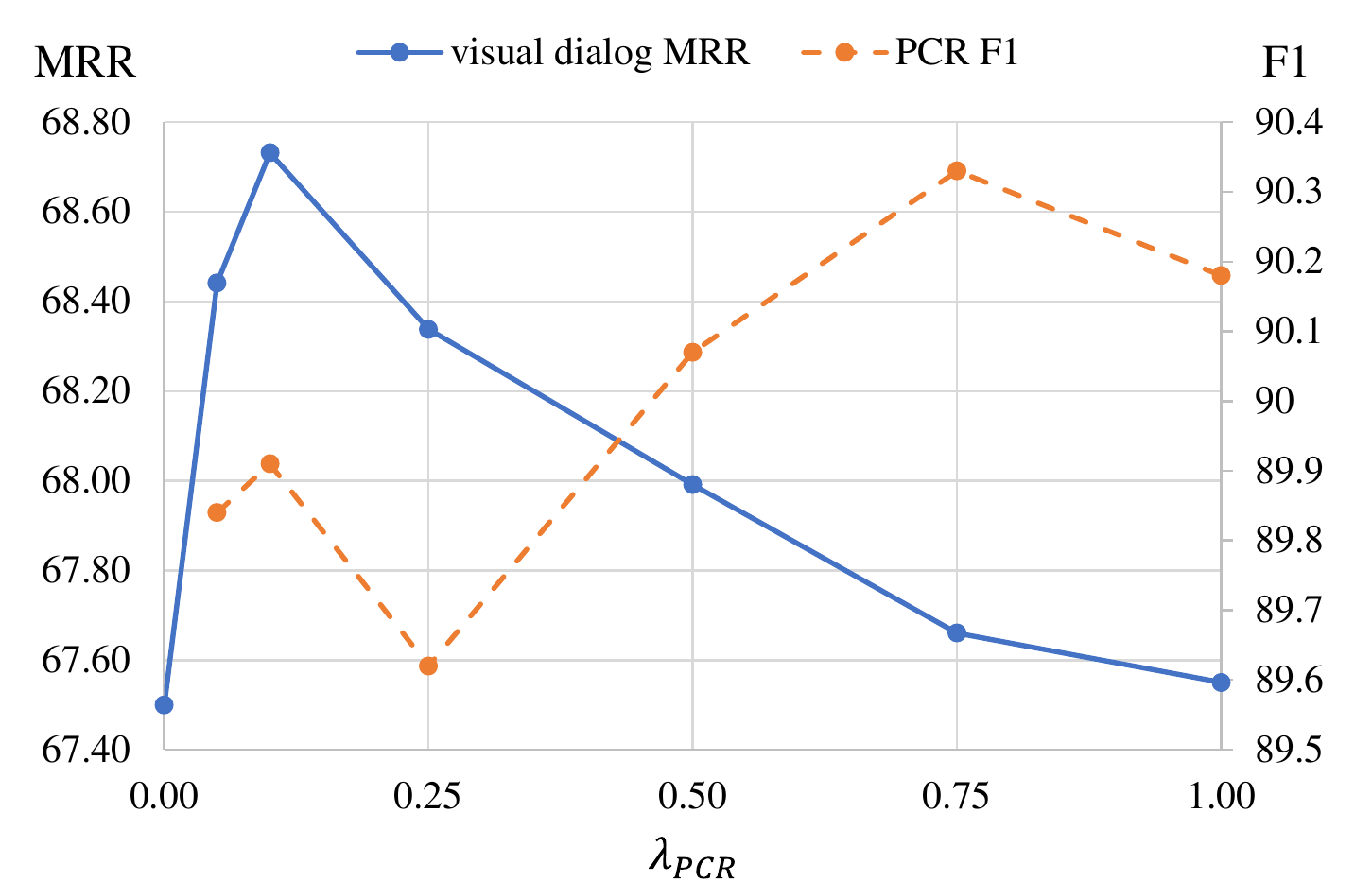}
  \caption{The effect of the loss efficient $\lambda_{PCR}$ on visual dialog and PCR metrics. Results are reported on the VisDial v1.0 test-standard set and the VisPro test set. When $\lambda_{PCR}=0$, the model does not learn from the PCR loss, so no F1 score is reported.}
  \label{fig:pcr_coeff}
\end{figure}

To understand the impact of the loss coefficients defined in Equation (\ref{eq:lambda}), we fix $\lambda_{NSP}$ and $\lambda_{MTM}$ to 1 following \cite{DBLP:conf/eccv/MurahariBPD20} and vary $\lambda_{PCR}$ from 0 to 1. As Figure~\ref{fig:pcr_coeff} shows, the MRR score of the visual dialog task first increases and then decreases, peaking at $\lambda_{PCR}=0.1$. For the F1 metric of the PCR task, larger $\lambda_{PCR}$ leads to higher F1. Results indicate that when $\lambda_{PCR}$ grows larger, the model tends to fit the PCR loss better and pay less attention to the visual dialog losses. When $\lambda_{PCR}$ is set properly, the joint training of PCR can help improve the visual dialog metrics.
A possible reason is that the visual dialog task is more complicated than the PCR task, so the visual dialog losses require larger coefficients.

\subsubsection{Results on questions with and without referential pronouns}
\begin{table}[t]
  \centering
  \caption{The difference of MRR scores between ViLBERT and our model with joint training and coreference-related heads on the VisDial v1.0 validation set. Questions are divided by whether they contain referential pronouns (denoted as w/ ref. prp.) or not (denoted as w/o ref. prp.).}
  \vspace{0.1in}
    \begin{tabular}{c|c|c|c|c}
    \toprule
    \multicolumn{1}{c}{\multirow{2}[4]{*}{question type}} & \multicolumn{1}{c}{\multirow{2}[4]{*}{\#questions}} & \multicolumn{2}{c}{MRR} & \multirow{2}[4]{*}{$\Delta$MRR} \\
\cmidrule{3-4}    \multicolumn{1}{c}{} & \multicolumn{1}{c}{} & \multicolumn{1}{c}{ViLBERT} & \multicolumn{1}{c}{MB-JC} &  \\
    \midrule
    w/ ref. prp. & 3809  & 68.22 & 69.21 & 0.99 \\
    w/o ref. prp. & 16791 & 69.30 & 69.99 & 0.69 \\
    \midrule
    all   & 20600 & 69.10 & 69.85 & 0.75 \\
    \bottomrule
    \end{tabular}%
  \label{tab:mrr_diff_question}%
  \vspace{-0.1in}
\end{table}%

We further compare the performance of models on questions with referential pronouns and those without in Table~\ref{tab:mrr_diff_question}. The increase by joint training is 0.99 on questions with referential pronouns and 0.69 on questions without.  It demonstrates that the jointly trained model achieves higher MRR because the PCR loss helps the model better interpret the referential pronouns in questions.

\subsubsection{Case study}

\begin{figure}[htbp]
  \centering
  \includegraphics[width=0.9\linewidth]{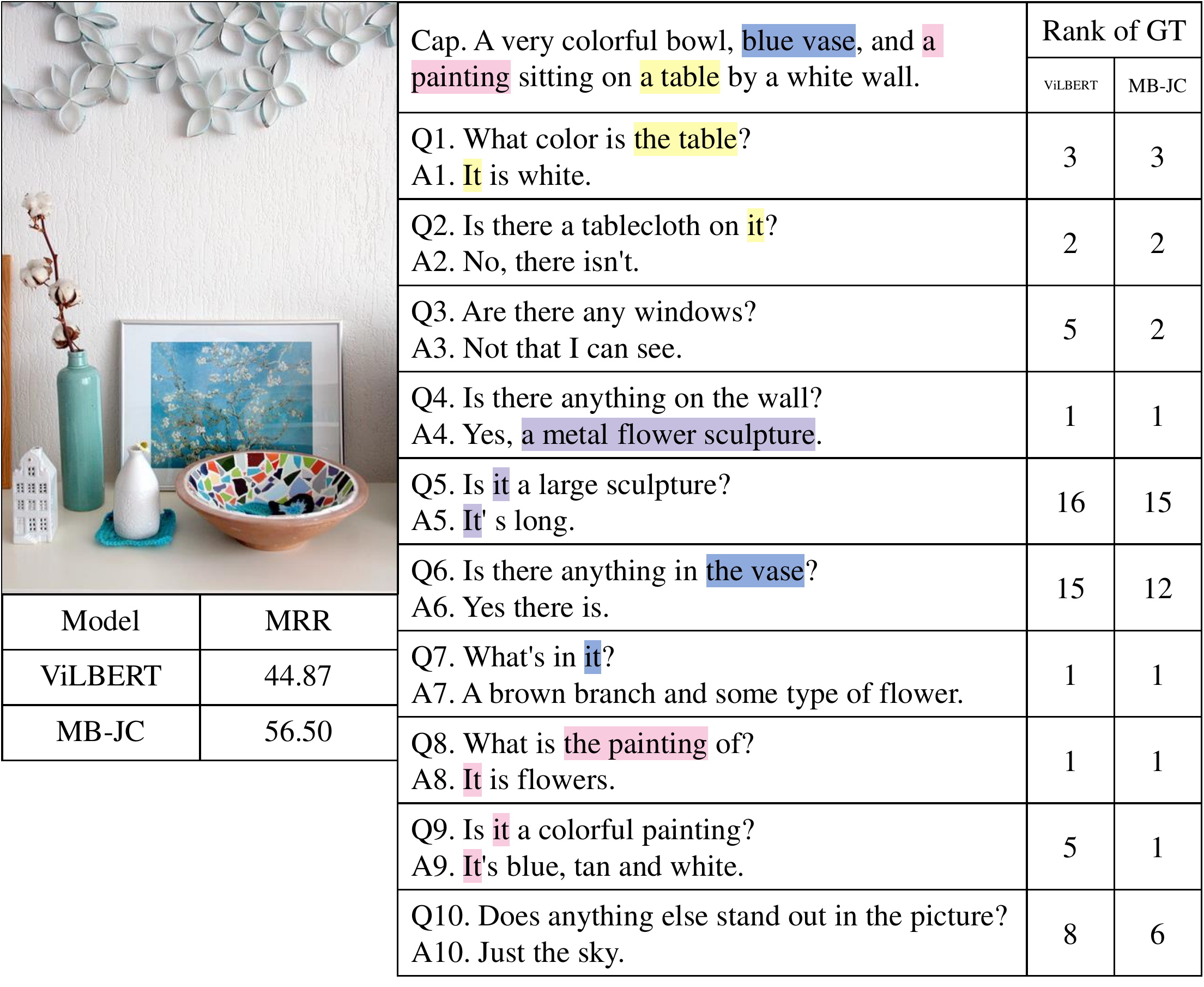}
  \caption{An example from the VisDial v1.0 validation set to compare ViLBERT and our model with joint training and coreference-related heads in Phase 1. Words in different coreference clusters are highlighted in different colors. For each round, the predicted rank of the ground-truth answer is reported.}
  \label{fig:mrr_case}
\end{figure}

We provide a specific case to analyze the effect of the proposed methods.
Figure~\ref{fig:mrr_case} presents predictions of different models on a dialog. Compared to ViLBERT, our jointly trained model can understand the pronouns better and thus rank the ground-truth answer higher. For instance, for $Q_9$ ``Is it a colorful painting?'' The model with joint training correctly identifies the ground-truth answer ``It's blue, tan and white'', while the ViLBERT ranks it in the $5^{th}$ place.

\subsection{Analysis of history pruning in Phase 2}

Table~\ref{tab:vd-main-1} shows that for single models in Phase 2, the previous best NDCG scores are achieved by ViLBERT at 74.47 and VD-BERT at 74.54. Our MB-JC-HP model with the \textit{crf + cap} rule achieves a higher NDCG at 75.30. For ensembled models, VD-BERT gets 75.35 NDCG. While ViLBERT~\cite{DBLP:conf/eccv/MurahariBPD20} reports no improvement from ensembling, our model that ensembles the outcome of models trained under four pruning rules boosts the performance from 75.30 to 76.14. The improvement for both single and ensembled models indicates the effectiveness of history pruning.

Compared to the MB-JC model for Phase 1, the MB-JC-HP model for Phase 2 increases NDCG from 63.55 to 75.30 while decreases MRR from 68.73 to 56.05. There are two reasons for such drop in retrieval metrics.
For one thing, as is discussed in Section~\ref{sec:visdial}, the ground-truth labels for retrieval and ranking metrics are misaligned and sometimes conflicting. In Table~\ref{tab:vd-main-1}, baseline models that crosses two phases such as ViLBERT and VD-BERT also experience similar improvement in NDCG and drop in retrieval metrics.
For another, as is explained in Section~\ref{sec:overview}, dialog history has different impacts in different phases. 
The overlap between a candidate answer and dialog history is useful for picking the best answer to maximize retrieval metrics but harmful for judging relevance for all answers to optimize ranking metrics.
Thus pruning dialog history favors ranking metrics but harms retrieval metrics.
Nonetheless, compared to the baseline ViLBERT model, our MB-JC-HP model achieves higher ranking scores as well as better  retrieval scores.

\subsubsection{Results of different strategies on different base models}

\begin{table}[t]
  \centering
  \caption{Results of single models trained under different rules for history pruning and their ensembles on the VisDial v1.0 test-standard set. Our VD-PCR model use MB-JC from Phase 1 as the base model. Best results are in \textbf{bold} font.}
  \vspace{0.1in}
    \begin{tabular}{c|p{1.3cm}<{\centering}|p{1.3cm}<{\centering}|p{1.3cm}<{\centering}|p{1.3cm}<{\centering}|c}
    \toprule
    \multicolumn{1}{c}{\multirow{2}[4]{*}{Base Model}} & \multicolumn{4}{c}{History Pruning Rule} & \multirow{2}[4]{*}{Ensemble} \\
\cmidrule{2-5}    \multicolumn{1}{c}{} & \multicolumn{1}{c}{\textit{crf + cap}} & \multicolumn{1}{c}{\textit{crf}} & \multicolumn{1}{c}{\textit{cap}} & \multicolumn{1}{c}{\textit{all}} &  \\
    \midrule
    MB-JC & 75.30 & 74.46 & 74.76 & 74.70 & \textbf{76.14} \\
    \midrule
    VD-BERT & 74.50 & 73.54 & 74.00 & 74.20 & \textbf{75.98} \\
    \bottomrule
    \end{tabular}%
  \label{tab:ensemble}%
  \vspace{-0.1in}
\end{table}%

To compare various pruning rules, we report the results of single models for each rule and their ensembles in Table~\ref{tab:ensemble}. 
As history pruning is easy to implement by only changing the input, we further apply our methods on VD-BERT~\cite{DBLP:conf/emnlp/WangJLKXH20}. For VD-BERT, we also observe an increase from 74.50 to 75.98 by ensembling. 
While the original VD-BERT~\cite{DBLP:conf/emnlp/WangJLKXH20} ensembles 12 models and improves NDCG with a margin of 0.81, our model ensembles only four models and increases NDCG by 1.48.

For both MB-JC and VD-BERT, the \textit{crf + cap} rule outperforms the \textit{all} rule with full history. It demonstrates that excluding irrelevant history helps the model rank all the candidate answers better. 
However, the \textit{crf} rule on VD-BERT with only the co-referring rounds performs worse than \textit{all}.

There are potentially two reasons for such a drop in performance.
First, as discussed in Section~\ref{sec:pruning_four}, the PCR model could make mistakes during prediction, so some relevant dialog history might be pruned mistakenly. Including the caption as part of the input alleviates the problem.
Second, the caption is the description of the image and the start of the dialog. 
It helps the multi-modal BERT model to align the two modalities, which is similar to the object tags to align vision and language in \cite{DBLP:conf/eccv/Li0LZHZWH0WCG20}.
These reasons also count for why the \textit{cap} rule that simply includes the caption gets comparable results with \textit{all}.

\subsubsection{Results on questions with and without referential pronouns}

\begin{table}[t]
  \centering
  \caption{The difference of NDCG scores between ViLBERT and our VD-PCR with different history pruning rules on the VisDial v1.0 validation set. Among 2,060 questions (denoted as ``total'') in the validation set, 367 questions contain referential pronouns (denoted as ``w/ ref. prp.'') and 1693 do not (denoted as ``w/o ref. prp.''). Best results are in \textbf{bold} font.}
  \vspace{0.1in}
    \begin{tabular}{c|c|l|l|l}
    \toprule
    \multicolumn{1}{c}{\multirow{2}[4]{*}{Model}} & \multicolumn{1}{c}{\multirow{2}[4]{*}{Rule}} & \multicolumn{3}{c}{Question Type} \\
\cmidrule{3-5}    \multicolumn{1}{c}{} & \multicolumn{1}{c}{} & \multicolumn{1}{c}{w/ ref. prp.} & \multicolumn{1}{c}{w/o ref. prp.} & \multicolumn{1}{c}{total} \\
    \midrule
    ViLBERT & \textit{all} & 74.69 & 75.65 & 75.47 \\
    \midrule
    \multirow{4}[2]{*}{VD-PCR} & \textit{crf + cap} & \textbf{75.34 (+0.65)} & \textbf{76.73 (+1.08)} & \textbf{76.47 (+1.00)} \\
          & \textit{crf} & 75.08 (+0.39) & 76.31 (+0.66) & 76.08 (+0.61) \\
          & \textit{cap} & 73.85 (-0.84) & 76.15 (+0.50) & 75.72 (+0.25) \\
          & \textit{all} & 75.00 (+0.31) & 75.85 (+0.20) & 75.70 (+0.23) \\
    \bottomrule
    \end{tabular}%
  \label{tab:ndcg_diff}%
  \vspace{-0.1in}
\end{table}%

Table~\ref{tab:ndcg_diff} shows the comparison of NDCG scores between ViLBERT and VD-PCR with different history pruning rules. 
All models get lower NDCG scores on questions with referential pronouns than questions without. It demonstrates that questions with pronouns are more complicated, consistent with the observation in Section~\ref{sec:mrr_clusters}.
For questions without referential pronouns, 
models with pruned input (\textit{crf + cap}, \textit{crf}, \textit{cap}) achieve better scores. It indicates that history pruning prevents the models from overfitting to unnecessary history contexts.
For questions with referential pronouns, the model with the \textit{cap} rule performs much worse than the baseline, 
while models with the \textit{crf + cap} and \textit{crf} rules still obtains higher scores. It demonstrates the effectiveness of pronoun coreference to identify relevant dialog history for question comprehension. The improvement on these questions is not as much as that on questions without pronouns. Although relevant history rounds are necessary for questions with pronouns, they could still distract the models with harmful bias.

\subsubsection{Case study}

\begin{figure}[t]
  \centering
  \includegraphics[width=0.9\linewidth]{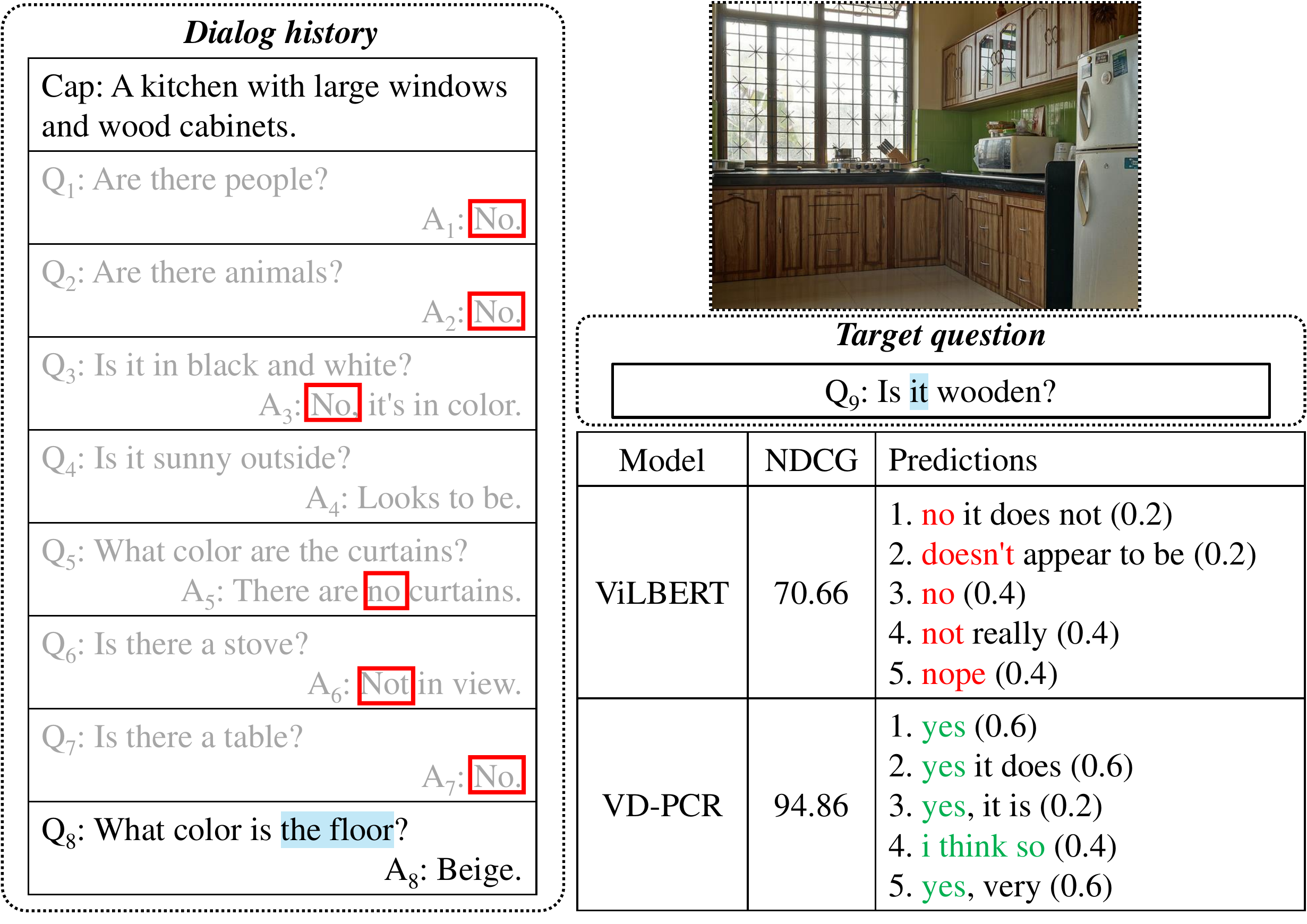}
  \vspace{-0.1in}
  \caption{An example from  the VisDial v1.0 validation set to compare the ViLBERT and our VD-PCR with history pruning. The models' top five predictions are listed with each prediction's annotated relevance score in the bracket. The referential pronoun ``it'' and its antecedent ``the floor'' are highlighted in blue. The dialog rounds in grey color are pruned from the input according to the rule \textit{crf + cap}.}
  \label{fig:ndcg_case}
  \vspace{-0.1in}
\end{figure}

In Figure~\ref{fig:ndcg_case}, we compare ViLBERT and VD-PCR on a dialog example. The ViLBERT with the entire dialog history as input is misled by ``no'' and ``not'' in the previous answers so it ranks negative answers at the top. The input of VD-PCR with the $crf + cap$ rule includes the caption and $(Q_8, A_8)$ as relevant rounds identified by coreference. Therefore, it avoids the distraction of irrelevant dialog history and ranks the positive responses as top answers.

\section{Conclusion}
\label{sec:conclusion}
In this paper, we propose a novel framework VD-PCR to improve visual dialog models with pronoun coreference.
First, to accurately resolve pronouns on VisDial, we design a PCR model by using a multi-modal BERT to fuse visual information and pseudo labels to utilize vast unlabelled data.
Then we leverage coreference in both implicit and explicit ways for VD-PCR.
To implicitly help models interpret pronouns in Phase 1, we use the joint training to simultaneously optimize the PCR and visual dialog tasks. While the visual dialog network takes the multi-modal BERT's last layer as token representations, the PCR network is based on the output of the selected coreference-related attention heads.
To avoid the negative impact of irrelevant history in Phase 2, we explicitly prune the irrelevant dialog history judged by pronoun coreference. 
With the help of pronoun coreference, VD-PCR achieves new state-of-the-art results on VisDial for both retrieval and ranking metrics of single and ensembled models.

Since understanding pronouns is a crucial step towards understanding language, the proposed methods to incorporate pronoun coreference could also be applied to other natural language applications to improve their performance. For joint training of high-level language understanding tasks and coreference on BERT, the latter can be solved on the coreference-related attention heads to guarantee compatibility between tasks. When dealing with long dialog history, pronoun coreference can be used to prune the irrelevant rounds and keep the key information. However, the proposed history pruning rules still have weaknesses such as neglecting ellipsis in dialogs. In future work, we will take ellipsis into account to design better history pruning criteria and apply our method to other dialog-related applications.

\bibliographystyle{plainnat}
\bibliography{main}

\end{document}